\documentclass[10pt,journal,compsoc]{IEEEtran}

\usepackage{framed,multirow}
\usepackage{amssymb}
\usepackage{latexsym}
\usepackage{graphicx}
\usepackage{multirow}
\usepackage{ulem}
\usepackage{color}

\usepackage{booktabs, tabularx, makecell, xcolor, threeparttable}
\usepackage{array}
\newcolumntype{Y}{>{\raggedright\arraybackslash}X} 
\definecolor{groupbg}{RGB}{255,235,238}
\newcommand{\modI}{\textbf{I}} 
\newcommand{\modV}{\textbf{V}} 
\newcommand{\modK}{\textbf{K}} 
\newcommand{\modE}{\textbf{E}} 

\newcommand{\groupheader}[1]{%
  \rowcolor{groupbg}\multicolumn{8}{c}{\textbf{#1}}\\
}

\usepackage[most]{tcolorbox}
\newenvironment{takeaways}{\begin{tcolorbox}[colback=black!3,colframe=black!20,title=Key takeaways]}{\end{tcolorbox}}

\usepackage[breaklinks=true,colorlinks,bookmarks=false,citecolor=green,linkcolor=green]{hyperref}
\usepackage[activate]{microtype}
\usepackage{svg}
\usepackage{natbib}
\usepackage{xcolor}
\usepackage{pifont} 

\usepackage{amsmath} 
\newcommand{\mathtext}[1]{\ifmmode\text{#1}\else#1\fi}

\newcommand{\xmark}{\mathtext{\textcolor{red!70!black}{\ding{55}}}}   

\usepackage{booktabs}

\usepackage{url}
\usepackage{colortbl}
\usepackage[separate-uncertainty = true,multi-part-units = repeat]{siunitx}
\usepackage{array,amsfonts,amsmath}
\usepackage{caption}
\usepackage{threeparttable}
\usepackage{subcaption}
\usepackage{rotating}
\usepackage{tablefootnote}
\usepackage[english]{babel}
\usepackage[utf8]{inputenc}
\usepackage{algorithm}
\usepackage{adjustbox}
\usepackage{afterpage}
\usepackage{sidecap}
\PassOptionsToPackage{noend}{algpseudocode}
\usepackage{algpseudocode}
\usepackage{pdflscape}
\usepackage{float,lscape}
\usepackage{scrhack}
\usepackage{epstopdf}

\usepackage[most]{tcolorbox}
\graphicspath{{./figures/}} 

\usepackage{boldline} 
\usepackage{footnotehyper}
\definecolor{mygray}{gray}{.9}

\begin{document}

\title{Surgical Scene Understanding in the Era of Foundation AI Models: A Comprehensive Review}

\author{Ufaq Khan$^*$, Umair Nawaz, Adnan Qayyum, Shazad Ashraf, Yutong Xie, Muhammad Haris Khan,\\ Muhammad Bilal, Junaid Qadir
\IEEEcompsocitemizethanks{\IEEEcompsocthanksitem  U. Khan, U. Nawaz, Y. Xie, M. H. Khan are with the MBZ University of AI, Abu Dhabi, UAE. \protect \\
$^*$Correspondence E-mail: ufaq.khan@mbzuai.ac.ae
\IEEEcompsocthanksitem A. Qayyum is with Hamad Bin Khalifa University (HBKU), Doha, Qatar.
\IEEEcompsocthanksitem S. Ashraf and M. Bilal are with Birmingham City University, United Kingdom.
\IEEEcompsocthanksitem J. Qadir is with Qatar University, Doha, Qatar.
\IEEEcompsocthanksitem S. Ashraf is also with the University Hospitals Birmingham and University of Birmingham, Birmingham, United Kingdom.
}
}



\IEEEtitleabstractindextext{%
\begin{abstract}
Recent advancements in machine learning (ML) and deep learning (DL), particularly through the introduction of Foundation Models (FMs), have significantly enhanced surgical scene understanding within minimally invasive surgery (MIS). This paper surveys the integration of state-of-the-art ML and DL technologies, including Convolutional Neural Networks (CNNs), Vision Transformers (ViTs), and Foundation Models like the Segment Anything Model (SAM), into surgical workflows. These technologies improve segmentation accuracy, instrument tracking, and phase recognition in surgical scene understanding. The paper explores the challenges these technologies face, such as data variability and computational demands, and discusses ethical considerations and integration hurdles in clinical settings. Highlighting the roles of FMs, we bridge the technological capabilities with clinical needs and outline future research directions to enhance the adaptability, efficiency, and ethical alignment of AI applications in surgery. Our findings suggest that substantial progress has been made; however, more focused efforts are required to achieve seamless integration of these technologies into clinical workflows, ensuring they complement surgical practice by enhancing precision, reducing risks, and optimizing patient outcomes.
\end{abstract}

\begin{IEEEkeywords}
Surgical Scene Understanding, Endoscopic Video Analysis, AI in Surgery, Future Directions in Surgical AI
\end{IEEEkeywords}}
\maketitle

\section{Introduction}
\label{s:intro}


Artificial intelligence (AI) has the potential to have a profound impact in the field of surgery. This is particularly relevant in the field of ``vision'', where improving the understanding of complex surgical scenes through advanced imaging and analysis techniques can complement surgical actions \citep{liu2024evolution}. Minimally invasive surgery (MIS) \citep{rivas2021review} has transformed modern operating rooms (OR) into visually intensive environments. Surgeons now primarily rely on real-time video feeds, guiding their movements through keyhole ports while interacting with surgical instruments. However, this reliance brings unique challenges that range from rapid camera movements, reflective glare, smoke, blood, and frequent occlusions that significantly degrade visual quality \citep{zhang2025uk}. Traditional and manually designed computer vision algorithms \citep{kitaguchi2022artificial,mascagni2022computer} struggle under these challenging conditions and often fail when visual scenarios deviate even slightly from predefined assumptions.

AI advancements such as deep learning (DL) have opened new possibilities for understanding surgical scenes. Unlike traditional methods \citep{kabir2025computer} that rely on predefined features, these AI models learn to recognize patterns and features directly from data, improving their ability to generalize across various settings. These capabilities are crucial for tasks such as distinguishing anatomical structures from surgical tools, navigating obscured views, and predicting the presence of abnormalities or complications in real time \citep{maier2017surgical}. Initially, DL approaches offered partial solutions by learning visual patterns from large annotated surgical datasets \citep{li2024deep,rodrigues2022surgical}. However, such early models frequently faltered in real-world settings, where visual contexts differed substantially from their training data. Addressing these limitations requires models capable of robust generalization and efficient adaptability to the dynamic and complex visual landscape of surgical scenes.

\begin{figure}[hbtp]
  \centering
  \includegraphics[width=\columnwidth]{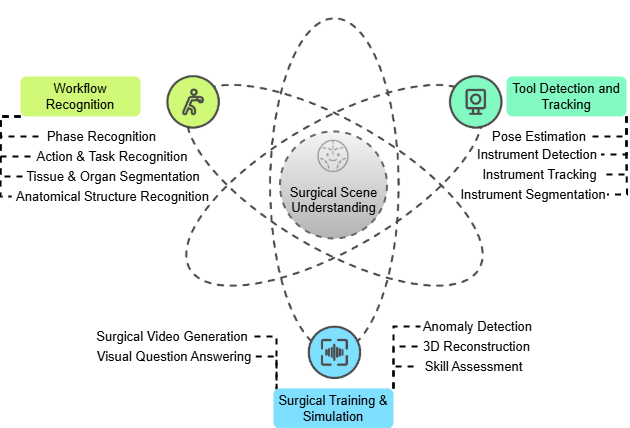}
  \caption{{Key components in surgical scene understanding: we explore the use of Surgical AI for \textit{Tool Detection and Tracking} in Section \ref{sec:tool_recognition}, \textit{Workflow Recognition} in Section \ref{sec:workflow} and \textit{Surgical Training and Simulation} in Section} \ref{sec:surgical_training}.}
  \label{fig:scene-understanding}
\end{figure}

\begin{figure*}[h!]
  \centering
  \includegraphics[width=\textwidth]{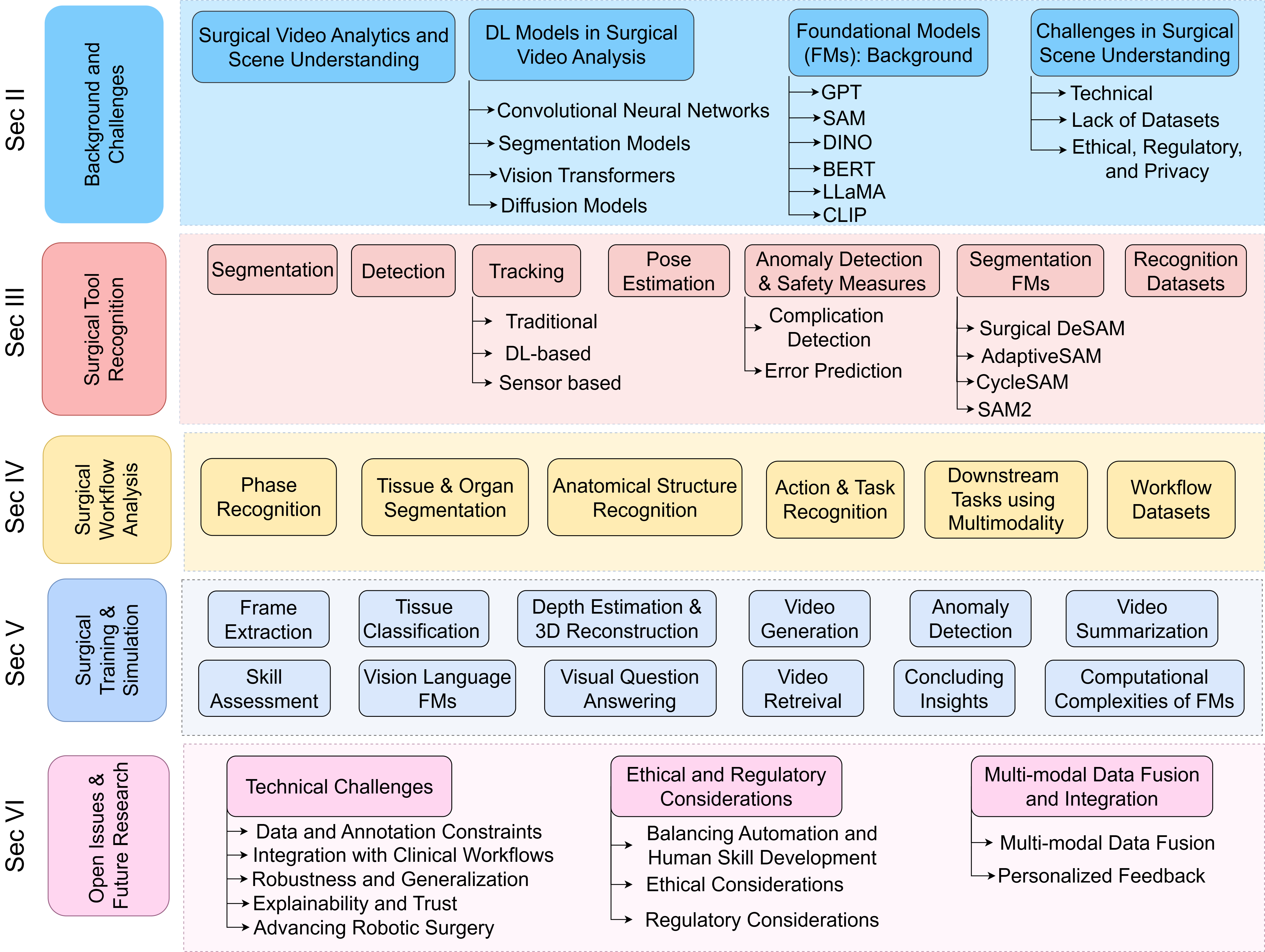}
  \caption{Organization of the review paper on surgical scene understanding categorized into seven sections, detailing the evolution of deep learning models in surgical applications, advancements in surgical tool recognition, and comprehensive analysis of surgical workflows, training, and simulation. The paper provides a structured roadmap through various technical discussions, highlights relevant datasets, and concludes with insights into open issues and future research directions in the domain of surgical scene understanding.}

  \label{fig:organization}
\end{figure*}

The advent of foundation models, such as the Segment Anything Model (SAM), DINO-v2, CLIP, and emerging video-language transformers \citep{kirillov2023segment,caron2021emerging,radford2021learning}, marks a transformative step in overcoming the aforementioned barriers. These large-scale, pre-trained networks utilize extensive and diverse datasets to build generalized representations, significantly enhancing their adaptability to unseen or challenging scenarios. Crucially, their ability to rapidly adapt to specific surgical tasks through lightweight fine-tuning methods, like prompt tuning \citep{yue2024surgicalsam} or adapter layers \citep{cui2024surgical,wei2024enhancing}, drastically reduces the annotation challenge, which is a longstanding bottleneck in medical AI.


These Foundation Models have been applied to multiple domain-specific applications \citep{chen2024overview,awais2025foundation,lu2025vision}, and that is why we investigate how this latest generation of foundational vision and multimodal models can be effectively specialized and applied to address key challenges in understanding the surgical scene. Our review focuses on three fundamental interconnected tasks that reflect real-world clinical needs, as illustrated in Figure~\ref{fig:scene-understanding}. 

Firstly, robust tool-and-object detection and tracking precisely identify and localize surgical instruments and anatomical structures, which is critical to improving surgical safety through context-aware warnings, facilitating robotic instrument assistance, and enabling augmented reality overlays \citep{tu2024head}. Secondly, workflow recognition systematically segments procedures into coherent phases and identifies surgeon gestures, which also directly supports automated documentation, progress tracking dashboards, and timely decision support systems. Thirdly, training and simulation harness advanced analytics, synthetic video generation, and anomaly detection to create realistic simulators and objective metrics to assess and refine surgical skills. Importantly, these tasks are not isolated, as each directly influences the effectiveness of the others. Accurate detection of surgical tools informs better recognition of surgical phases, while a precise workflow context enhances the realism and educational value of simulated training materials. Foundation models serve as a critical integrative mechanism by enabling shared representations across these tasks, reinforcing their collective accuracy and applicability.

\subsection{Comparison with Related Surveys}
Despite significant recent advances, the existing literature remains fragmented. DL applications in medical imaging and surgery have been extensively reviewed, with numerous studies summarizing advancements and identifying emerging trends. Previous works, such as those of \citep{rivas2021review} and  \citep{rueckert2024methods}, have predominantly focused on specific tasks such as automation in MIS and instrument segmentation, respectively. In contrast, this survey expands the focus to encompass a broader spectrum of challenges, including segmentation, tracking, and workflow recognition. Moreover, unlike the studies by \citep{fu2021future} and  \citep{azad2024medical} that mainly focuses on endoscopic navigation and U-Net variants, respectively, and explore individual methodologies, this paper emphasizes the applicability of advanced models such as Vision Transformers (ViTs), Large Vision-Language Models (LVLMs), and foundation models like SAM. These models are particularly highlighted for their potential to assist and enable real-time decision making and address the complexities associated with multimodal data. Furthermore, based on the work of \citep{li2024deep}, this survey enhances existing dataset analyses and presents an updated and detailed catalog specifically tailored for surgical scene understanding tasks. By synthesizing these contributions, this survey not only consolidates current knowledge but also identifies key trends, such as the integration of multimodal AI and foundation models for surgical applications. It also underscores the pressing need for real-time decision-making tools that meet the dynamic and  ``high-stake'' demands of live surgical environments, providing a robust framework to advance AI-driven innovations in surgical practice.


In Table \ref{tab:comparisonrelated}, we present a comprehensive comparison of this article with existing surveys and review articles that have a similar focus but leave a substantial gap in synthesizing broader trends towards foundation model approaches. Specifically, a comparison is provided in terms of scope, procedural coverage, datasets utilized, methodological approaches, algorithmic details, performance metrics, and unique contributions. Compared to the existing literature, this survey adopts a more comprehensive and integrative approach and attempts to address various methodologies and applications within the domain of understanding the surgical scene. It explicitly bridges this gap, providing a comprehensive review that examines in parallel the benefits of CNN and transformer-based vision models for detection in challenging visual conditions, the effectiveness of video-language transformers in achieving near zero-shot recognition of surgical phases, and the use of diffusion-based models for realistic and customized training simulations.

\begin{figure*}[hbtp]
    \centering
    \includegraphics[width=1\textwidth]{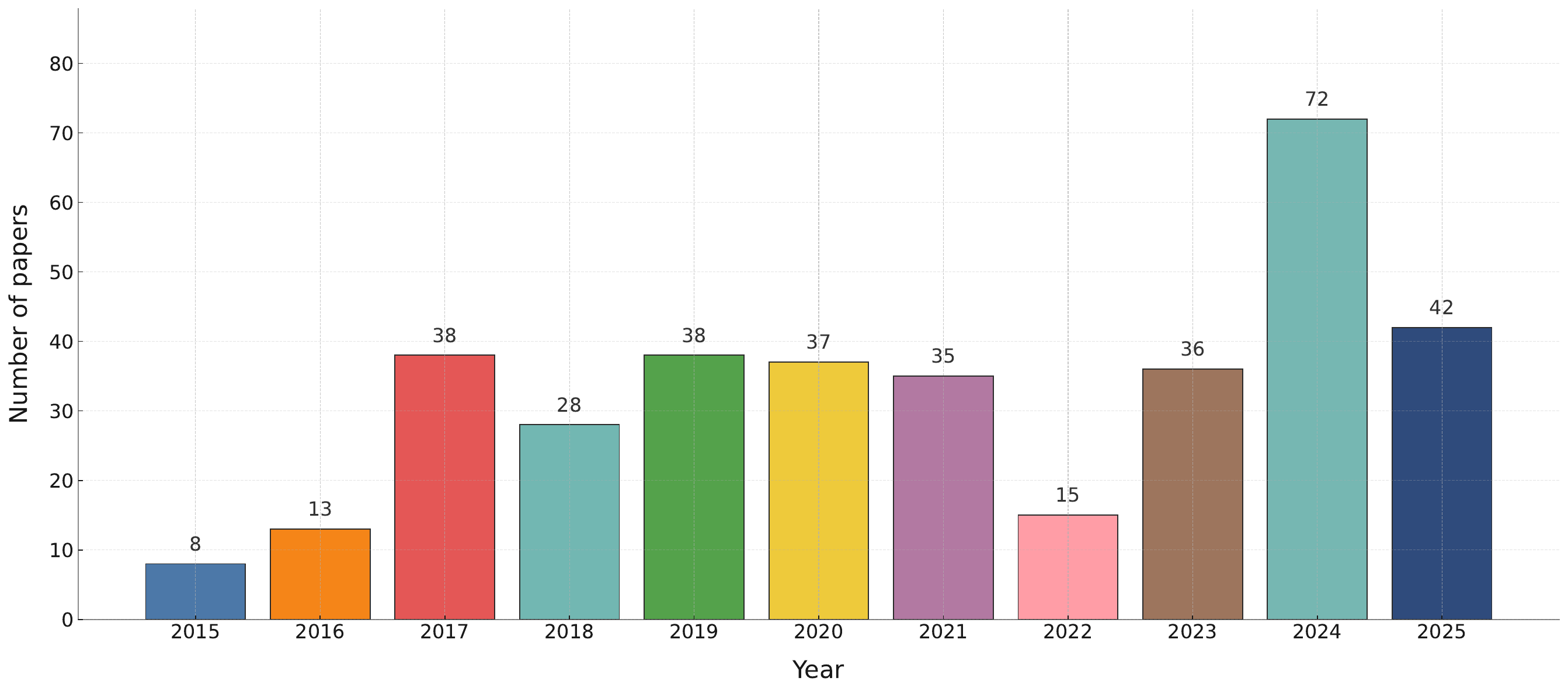}
    \caption{Papers per year in surgical scene understanding (2015–2025) drawn from this survey’s reference list. The trajectory highlights accelerated activity since 2017 with a peak around 2024. Here, the 2025 value is author-curated as of Aug 31, 2025 (n=42). Tasks include instrument/anatomy segmentation, instrument detection, workflow/phase recognition, and action triplets in laparoscopic/robotic/endoscopic videos.}
    \label{fig:survey-paper-count}
\end{figure*}




\subsection{Organization of This Paper}
Throughout this comprehensive review, we highlight persistent challenges such as domain shifts, stringent real-time constraints, privacy concerns, and annotation scarcity. Furthermore, we discuss emerging solutions enabled by scalable pre-training and efficient fine-tuning, detailing how these techniques provide pathways towards overcoming existing barriers.

Figure~\ref{fig:organization} outlines the structure of our survey, guiding the readers through its systematic exploration. Section~\ref{sec:back} introduces key concepts in surgical video analytics, discussing domain-specific challenges, including data scarcity, class imbalance, and ethical considerations. Section~\ref{sec:tool_recognition} covers detection, segmentation, and tracking methods, tracing the evolution from classical YOLO-inspired architectures to advanced transformer-based models and promptable frameworks like SAM. Section~\ref{sec:workflow} discusses workflow recognition, highlighting temporal and multimodal methodologies, and showcasing how video-language transformers and contrastive learning significantly outperform conventional recurrent architectures. Section~\ref{sec:surgical_training} focuses on training and simulation, examining the capabilities of diffusion models for synthetic data generation, interactive question-answering systems based on vision-language models, and skill assessment frameworks leveraging contrastive learning. Section~\ref{sec:open-issues} identifies critical open issues and outlines promising future research directions, including federated learning approaches, domain-adaptive fine-tuning methods, regulatory considerations, and the urgent need for explainability. Finally, Section~\ref{sec:concs} concludes the discussion, summarizing key insights and strengthening the critical role that foundation models play in shaping the future of understanding the surgical scene.




\subsection{Contributions of This Paper}
In summary, this survey contributes uniquely by presenting: 

\begin{enumerate}
    \item \textit{Contemporary Focus on Foundation AI Models:} This survey uniquely highlights how foundation models are reshaping surgical scene understanding, particularly through their ability to generalize across tasks and modalities. We analyze their applications in real-time surgical workflows for tasks such as robust tool segmentation, fine-grained workflow recognition, and anomaly detection, emphasizing their scalability and adaptability in MIS. Figure~\ref{fig:survey-paper-count} shows the number of papers included in our bibliography by publication year (2015–2025).
    

    \item \textit{Critical Insights into Transformative Use Cases:} A detailed comparison of legacy CNN-based methods and contemporary transformer-based architectures, clarifying where foundation models currently excel.
    By focusing on endoscopic video analysis, we underscore how foundation models outperform traditional approaches in addressing challenges like variability in surgical scenes, occlusions, and inter-patient heterogeneity. 

    \item \textit{Evaluation of Evolving Datasets and Benchmarks:} We offer an in-depth evaluation of emerging datasets and benchmarks specifically designed for foundation model training and validation in surgical applications. This includes an analysis of large-scale, multimodal datasets that enable transfer learning and fine-tuning for surgical scene understanding, addressing gaps in data availability and variability. practical recommendations for data curation, fine-tuning methodologies, evaluation protocols, and clinical deployment strategies, offering researchers, engineers, and clinicians a clear, actionable roadmap.
\end{enumerate}

By centering on foundation AI models, this survey represents a timely and significant advancement in the literature, offering a forward-looking perspective on their role in revolutionizing surgical practices. It serves as a critical resource for researchers and clinicians aiming to leverage state-of-the-art AI to enhance surgical scene understanding in MIS, a fundamental step in improving patient outcomes.

\

\begin{figure*}[hbtp]
  \centering
  \includegraphics[width=0.7\textwidth]{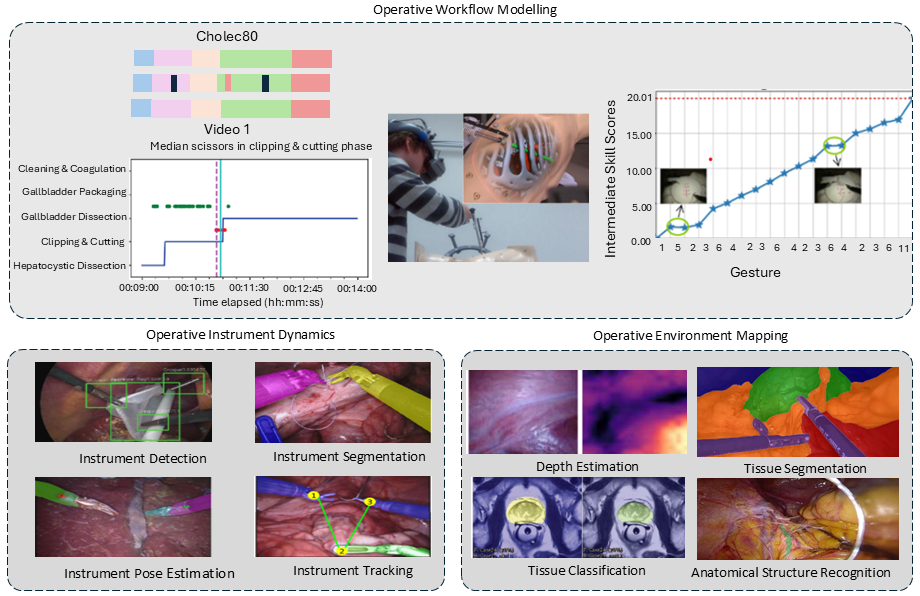}
  \caption{{An in-depth examination of the intricate tasks involved in navigating and controlling surgical tools, highlighting the technical and operational processes that enable precision and effectiveness in modern surgical procedures}}
  \label{fig:tool-centric}
\end{figure*}

\begin{table*}[!h]
\centering
\caption{Comparative performance and limitations of DL-based medical image segmentation surveys.}
\label{tab:comparisonrelated}
\resizebox{\textwidth}{!}{%
\begin{tabular}{@{}p{1.8cm}p{2.5cm}p{2.5cm}p{2cm}p{2cm}p{2cm}p{1.7cm}p{1.8cm}p{1.5cm}p{2cm}p{2cm}p{2cm}p{1cm}@{}}
\toprule
\textbf{Reference} & \textbf{Scope} & \textbf{Contributions} & \textbf{Algorithms} & \textbf{Generative Models} & \textbf{Vision Language Models} & \textbf{Foundation Models} & \textbf{Ethical and Regulatory Insights} & \textbf{Dataset Coverage} & \textbf{Comprehensive Surgical Tasks} & \textbf{Foundation Models Adaptation} & \textbf{Clinical Perspective} & \textbf{Open Issues} \\ \midrule

Rivas-Blanco et al. \citep{rivas2021review} & DL models in minimally invasive surgery & Deep insights into automation of surgical tasks & CNN, RNN, ANN, HMM & \centering\xmark & \centering\xmark & \centering\xmark & \centering\xmark & \centering\checkmark & \centering\xmark & \centering\xmark & \centering\xmark & \xmark \\

Fu et al. \citep{fu2021future} & Endoscopic navigation technologies & Detailed exploration of endoscopic vision technologies & Augmented reality, various imaging modalities & \centering\xmark & \centering\xmark & \centering\xmark & \centering\xmark & \centering\xmark & \centering\xmark & \centering\xmark & \centering\checkmark & \xmark \\

Rueckert et al. \citep{rueckert2024methods} & Instrument segmentation in surgery & Comprehensive review on surgical instrument segmentation & Deep learning, CNN & \centering\xmark & \centering\xmark & \centering\xmark & \centering\xmark & \centering\checkmark & \centering\checkmark & \centering\xmark & \centering\xmark & \xmark \\

Zhang et al. \citep{zhang2024segment} & SAM in medical image segmentation & Exploration of SAM's extension to medical segmentation & SAM, foundation models & \centering\xmark & \centering\xmark & \centering\checkmark & \centering\xmark & \centering\xmark & \centering\xmark & \centering\checkmark & \centering\checkmark & \checkmark \\

Azad et al. \citep{azad2024medical} & Evolution and success of U-Net in medical imaging & In-depth analysis of U-Net variants across modalities & U-Net and its variants & \centering\xmark & \centering\xmark & \centering\xmark & \centering\xmark & \centering\checkmark & \centering\xmark & \centering\xmark & \centering\checkmark & \checkmark \\

Schmidt et al. \citep{schmidt2024tracking} & Tracking and mapping in medical CV & Insightful review of tracking/mapping in deformable tissues & Nonrigid tracking, SLAM & \centering\xmark & \centering\xmark & \centering\xmark & \centering\xmark & \centering\checkmark & \centering\xmark & \centering\xmark & \centering\checkmark & \checkmark \\

Upadhyay et al. \citep{upadhyay2024advances} & DL methods to overcome data scarcity & Extensive review on overcoming data scarcity & CNN, U-Net, GAN & \centering\xmark & \centering\xmark & \centering\xmark & \centering\checkmark & \centering\checkmark & \centering\xmark & \centering\xmark & \centering\xmark & \xmark \\

Zhou et al. \citep{zhou2020application} & AI in surgery integration & Comprehensive integration of AI in surgery & CNN, RNN & \centering\xmark & \centering\xmark & \centering\xmark & \centering\xmark & \centering\checkmark & \centering\xmark & \centering\xmark & \centering\checkmark & \checkmark \\

Morris et al. \citep{morris2023deep} & DL applications in sub-specialties & Insights into practical DL applications in surgery & CNN, RNN & \centering\xmark & \centering\xmark & \centering\xmark & \centering\xmark & \centering\xmark & \centering\xmark & \centering\xmark & \centering\checkmark & \xmark \\

Li et al. \citep{li2024deep} & Surgical workflow recognition & Detailed surgical workflow recognition analysis & CNN, LSTM & \centering\xmark & \centering\xmark & \centering\xmark & \centering\checkmark & \centering\checkmark & \centering\xmark & \centering\xmark & \centering\checkmark & \checkmark \\

Garrow et al. \citep{garrow2021machine} & Automated phase recognition & Highlight on phase recognition automation in surgery & HMM, ANN & \centering\xmark & \centering\xmark & \centering\xmark & \centering\xmark & \centering\checkmark & \centering\xmark & \centering\xmark & \centering\xmark & \xmark \\

This Paper & Surgical Scene Understanding & Detailed model review, applications overview, dataset analysis & CNNs, ViTs, LVLMs, GANs, FMs & \centering\checkmark & \centering\checkmark & \centering\checkmark & \centering\checkmark & \centering\checkmark & \centering\checkmark & \centering\checkmark & \centering\checkmark & \checkmark \\

\bottomrule
\end{tabular}
}
\end{table*}

\section{Background and Challenges}
\label{sec:back}

\subsection{Surgical Video Analytics and Scene Understanding}  
Surgical video analytics and scene understanding represent transformative advances at the intersection of computer vision, ML, and surgical practice, with the aim of improving the safety, efficiency, and precision of surgical procedures. These technologies leverage cutting-edge techniques from image processing and ML / DL to analyze videos captured during surgery, providing actionable insights in real time and impact on postoperative events (for example, increased bleeding events may lead to increased length of hospital stay; increased intraoperative bowel handling may lead to bowel paralysis and higher rates of post-operative nausea and vomiting). Algorithms for detecting, segmenting, and tracking surgical instruments and maneuvers transform complex video data into precise information that supports decision making, as illustrated in Fig.~\ref{fig:tool-centric}.

\begin{figure*}[h]
  \centering
  \includegraphics[width=\textwidth]{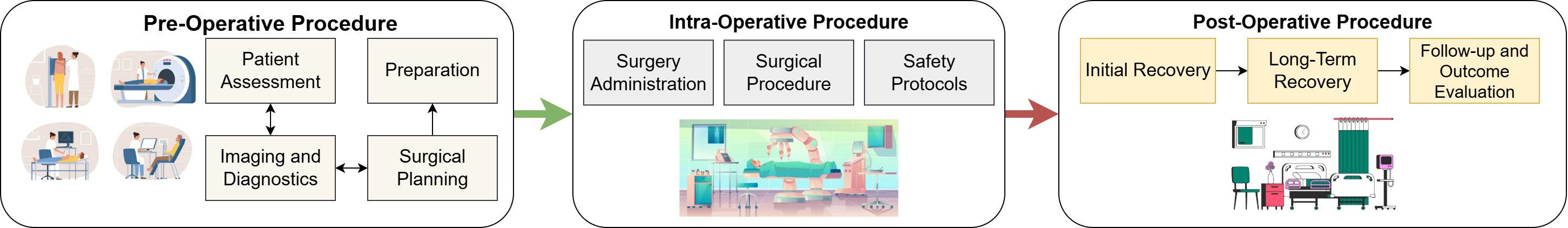}
  \caption{Sequential stages of a surgical procedure. {The \textit{Pre-Operative stage} involves patient assessment, detailed imaging, and surgical planning, which occur in a loop or with the possibility of back and forth. The \textit{Intra-Operative stage} focuses on the actual surgical procedure itself, including surgical administration, which includes monitoring, management, correct patient positioning, and adherence to safety protocols. The \textit{Post-Operative stage} covers both initial and long-term recovery, concluding with follow-up and outcome evaluations to ensure the effectiveness of the procedure.}}
  \label{fig:intraoperative}
\end{figure*}

Surgical scene understanding complements video analytics by focusing on the interpretation of surgical environments, including instrument recognition, workflow analysis, and scene segmentation. These capabilities enable real-time decision support through augmented overlays that guide procedural steps, comprehensive postoperative analyzes to assess surgical efficacy, and automated documentation to reduce administrative burden on surgical teams, particularly relevant after a procedure that may take several hours. The latter would allow clinicians to focus on other clinical tasks such as informing relatives of patients about the procedure or assisting in the safe transfer of patients out of the operating room. In addition, it is likely to describe the ground truth of operative events without bias, a highly relevant point in governance. In Fig.~\ref{fig:intraoperative}, the three stages of surgical procedures are depicted, illustrating the comprehensive workflow from pre-operative planning through intra-operative execution to post-operative recovery and evaluation.

Furthermore, the advent of 3D augmentation technologies has revolutionized surgical video analytics by converting traditional 2D video streams into dynamic 3D reconstructions, providing surgeons with an immersive and intuitive view of the surgical site \citep{yuk2021current}. These systems improve spatial awareness, simplify navigation around critical anatomical structures, and enable more precise interventions \citep{kim2015design}, minimizing patient trauma and improving outcomes.

Despite these advances, numerous challenges persist, including variability in surgical procedures, differences in patient anatomy, and the unstructured nature of surgical environments, which collectively complicate algorithm development. Moreover, high-dimensional video data demands substantial computational resources for real-time processing, while stringent privacy and security measures remain critical to safeguarding sensitive medical footage. The future of surgical video analytics and scene understanding lies in deeper integration with AI and multimodal data, combining video analysis with patient vitals and other intraoperative information. Innovations like edge computing promise real-time analytics in resource-constrained settings, whereas enhanced algorithms can provide greater granularity and accuracy in data interpretation. These advancements are poised to redefine surgical precision, training, and patient care, ultimately paving the way for safer and more effective surgical practices worldwide.

\subsection{DL Models in Surgical Video Analysis}
DL has substantially advanced computer vision, delivering remarkable accuracy in complex visual tasks by learning hierarchical features directly from large datasets. Surgical video analysis has greatly benefited from DL, as traditional handcrafted image processing methods struggle to generalize under the dynamic and visually challenging conditions of surgery \citep{hussain2022deep}. In particular, DL models have enabled breakthroughs in interpreting surgical scenes by accurately localizing tools, segmenting anatomical structures, recognizing procedural phases, and synthesizing realistic surgical scenarios. 

\subsubsection{Convolutional Neural Networks}
CNNs are fundamental to modern video analysis, particularly due to their ability to efficiently extract high-level features from visual data. Structured as a series of convolutional layers, CNNs capture spatial hierarchies in images, rendering them extremely effective in tasks such as object detection and scene classification \citep{bamba2021object}. These networks perform convolutions across image pixels to generate feature maps that summarize the presence of specific features at various locations in the image. This ability makes CNNs particularly adept at processing visual inputs \citep{khan2023pd} that are common in video data, such as frames from surgical procedures. In surgical video analysis, CNNs are utilized not only for object identification but also for understanding the interaction and relative positioning of various surgical tools and anatomical structures \citep{ahmed2024deep}. For example, through frame-by-frame analysis of video data, CNNs can detect environmental changes, track the movement of surgical instruments, monitor the progress of surgical interventions and any potential hazards, and predict the end of the surgical procedure (this helps in high-level management of theater lists and helps improve case throughput and theater time utilization). Such capabilities are crucial for developing real-time feedback systems that can alert surgeons about critical events or anomalies detected during surgery, as well as providing theater teams with a high-level overview of surgical procedures \citep{garcia2017toolnet}.

In medical image analysis, segmentation is a crucial task that provides detailed pixel-level annotations that are essential for both diagnostic and interventional procedures. Two of the most influential segmentation models in this domain are U-Net \citep{ronneberger2015u} and DeepLabv3 \citep{chen2017rethinking}, each known for its effectiveness in handling the complexities of medical imagery.
\begin{itemize}
    \item {\textit{U-Net}}: Developed specifically for biomedical image segmentation, U-Net features a symmetric architecture with a downsampling path to capture context and a precisely corresponding upsampling path to allow for precise localization \citep{ronneberger2015u}. This network architecture is particularly well-suited for medical applications because of its efficiency in using a limited number of training samples to produce high-resolution segmentations. The ability of U-Net to perform well even with small amounts of data and its robustness against noise in images make it a preferred choice for segmenting surgical video frames, where precise delineation of organs and surgical instruments is critical.
    \item {\textit{DeepLabv3}}: Building on earlier versions, DeepLabv3 incorporates \textit{atrous convolutions}, a technique that increases the receptive field of filters by inserting spaces between kernel elements, to capture richer contextual information without compromising the sharpness of the image segmentation \citep{chen2017rethinking}. This model utilizes an atrous spatial pyramid pooling module to robustly segment objects at multiple scales, a feature particularly useful for the varied scales seen in surgical videos. DeepLabv3's ability to effectively segment fine structures at different depths and scales is invaluable in surgeries, providing clear delineations that can guide surgical interventions and improve outcomes. 
\end{itemize}
In addition to the widely adopted U-Net and DeepLab variants, recent work has introduced architectures that combine the strengths of convolutional feature extraction with attention-based decoding or transformer-inspired meta-architectures. For example, Efficient Multi-scale Convolutional Attention Decoding (EMCAD) \citep{rahman2024emcad}) enhances encoder–decoder segmentation by introducing a lightweight, multi-scale attention decoding block. This design improves contextual feature aggregation across multiple resolutions while maintaining computational efficiency, making it well-suited for real-time surgical settings where GPU resources are constrained. Similarly, RAPUNet \citep{lee2024foundation} integrates a MetaFormer backbone with CNN layers for robust medical image segmentation, originally targeting polyp detection but with clear potential in surgical scene understanding. Its hybrid design leverages transformer-style token mixing for global context and CNN-based local feature extraction, enabling accurate segmentation of small and occluded structures. 

\begin{figure*}[hbtp]
  \centering
  \includegraphics[scale=0.55]{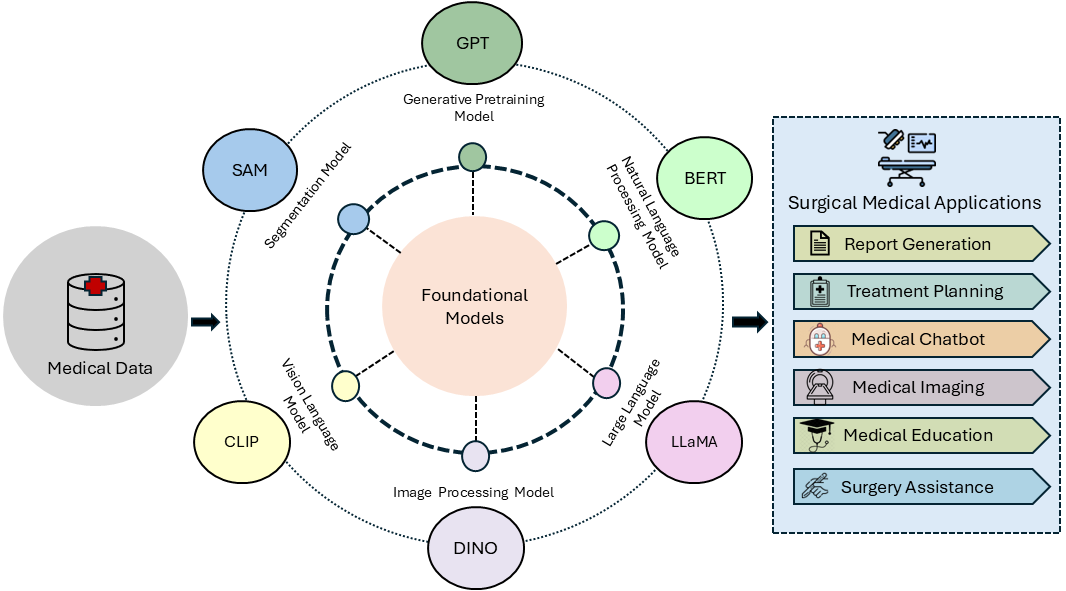}
  \caption{{Overview of Foundation Models applied in surgical settings, illustrating their roles in various applications such as report generation, treatment planning, and surgical assistance.}}
  \label{fig:foundational}
\end{figure*}


\subsubsection{Vision Transformers (ViT)}
ViT represents a significant paradigm shift in how image data is processed for complex tasks like video analysis, including in high-stakes environments such as operating theaters. Originally adapted from the transformer architecture, which has revolutionized natural language processing, ViTs apply the principles of self-attention to visual contexts, allowing them to learn contextual relationships between different parts of an image \citep{dosovitskiy2020image}. In surgical video analysis, ViTs have shown great promise in accurately segmenting and identifying critical structures within surgical scenes, outperforming traditional CNNs in tasks that require understanding complex spatial dependencies and long-range interactions \citep{carion2020end}. This makes ViTs particularly suitable for applications where precision and context-aware decision-making are crucial, such as real-time surgical guidance and post-operative analysis.

\subsubsection{Diffusion Models}
Diffusion models are a class of generative models that have attracted significant attention because of their ability to generate high-quality, detailed images from a learned distribution of training data. These models work by initially learning to gradually add noise to an image, transforming it into a Gaussian noise distribution, and then learning to reverse this process to reconstruct the original image from the noise \citep{ho2020denoising}. This forward and reverse process enables diffusion models to effectively model the probability distribution of training data, making them extremely powerful for generating or reconstructing images. 

In surgical video analysis, diffusion models can be utilized to enhance low-resolution or noisy surgical videos, providing clearer visualizations of surgical procedures \citep{cho2024surgen}. They can also generate synthetic surgical images for training purposes, allowing the creation of diverse scenarios that help train surgical staff without the need for extensive real-life video collections. Furthermore, diffusion models, with their generative capabilities, have significant potential to simulate possible surgical outcomes based on intraoperative videos, which helps surgeons plan and make decisions during complex procedures \citep{rodler2024generative}. These applications demonstrate the versatility and potential of diffusion models in transforming surgical education for future generations of surgical trainees and practice by enhancing the quality and utility of surgical imagery. Ultimately, a library of surgical videos that creates all the potential scenarios or hazards that can occur during surgery will lead to better surgeons and improved patient outcomes. 

Recent advances in diffusion-based approaches have shown strong potential for surgical video generation and planning. HieraSurg \citep{biagini2025hierasurg} introduces a hierarchy-aware framework that models surgical scenes at multiple abstraction levels. It employs a two-stage pipeline where a Semantic-to-Map module predicts future panoptic segmentation maps from the current frame, surgical phase, and action triplets, followed by a Map-to-Video module that synthesizes realistic videos from these maps. To overcome annotation scarcity, an automated segmentation pipeline based on SAM2 is used. This design enables HieraSurg to capture both coarse semantic changes and fine-grained textures, achieving superior visual quality, temporal coherence, and adherence to surgical content compared to prior methods.

To further enhance controllability, SurgSora \citep{chen2024surgsora} integrates object-aware motion guidance into the diffusion process. Instead of depending solely on RGB inputs or pre-defined segmentation masks, it combines RGB and depth features with segmentation cues through a Dual Semantic Injector for improved spatial understanding. A Decoupled Flow Mapper fuses these features with optical flow for realistic motion modeling, while a Trajectory Controller allows users to define precise tool or tissue movements via simple click-based inputs. Built upon Stable Video Diffusion, SurgSora generates high-fidelity videos that expert surgeons found nearly indistinguishable from real footage, while enabling fine-grained control over motion trajectories.

Beyond video synthesis, diffusion models have also been explored for surgical procedure planning. Multi-Scale Phase-Condition Diffusion (MS-PCD) \citep{zhao2024see} addresses the problem of predicting a sequence of actions to transition from a current surgical scene to a desired visual goal. The framework first performs phase recognition and then uses a multi-scale cascaded diffusion process conditioned on the phase and visual features at different spatial scales. By adaptively selecting the optimal input scale, MS-PCD captures subtle visual cues and improves action plan accuracy. Evaluated on the PSI-AVA dataset, it achieves notable gains over existing planning methods, demonstrating the adaptability of diffusion models to high-level decision-making tasks in surgical contexts.

\subsection{Foundation Models}

Foundation Models (FMs) represent a distinct class of machine learning models that are trained at scale on massive, heterogeneous datasets and subsequently adapted for a wide variety of downstream tasks \citep{10.1093/bjs/znae090}. Unlike conventional ML or even task-specific DL architectures, FMs are designed to capture broad, transferable representations that can be specialized through fine-tuning or prompting. This paradigm has already reshaped many domains, and researchers are increasingly investigating its impact on medical imaging and surgical applications. 

In medical imaging, FMs have demonstrated remarkable capabilities for interpreting complex visual data. Their large-scale pre-training allows them to learn nuanced feature hierarchies, which are critical for tasks such as segmentation, classification, and anomaly detection. Early studies show that FMs adapted to medical contexts can improve the accuracy and efficiency of identifying and delineating anatomical structures and pathological conditions \citep{khan2025comprehensive,shi2024survey,sun2024medical,zhang2024segment}. This represents a shift from narrow, handcrafted pipelines toward generalizable models that can be rapidly customized for different imaging modalities and clinical tasks.

Despite these benefits, several challenges limit the direct deployment of FMs in healthcare. First, data privacy concerns restrict access to diverse and sufficiently large medical datasets \citep{zhang2022privacy}. Second, the ``black box'' nature of large DL models creates difficulties in interpretability and accountability, which are common issues and a particular concern in safety and critical clinical decision-making \citep{rudin2019stop}. Third, many hospitals lack the computational resources or infrastructure required to fine-tune and deploy such large models \citep{topol2019high}. These challenges are compounded by strict regulatory frameworks such as HIPAA in the United States and GDPR in Europe, which impose further restrictions on data sharing and model development.

Nevertheless, FMs open up new possibilities for the medical domain. Their ability to integrate multimodal data, such as combining imaging with electronic health records, offers more comprehensive and context-sensitive diagnostic support \citep{zhang2024data,liu2024visual}. In surgical contexts, FMs show promise for video analysis \citep{schmidgall2024general}, anomaly detection \citep{kondepudi2025foundation}, and workflow optimization, where large-scale pre-training helps overcome the scarcity of annotated surgical datasets. These models also introduce opportunities for cross-institutional generalization, enabling methods trained in one type of procedure or population of patients to transfer effectively to another. At the same time, adapting FM for clinical practice requires domain-specific fine-tuning, interpretability mechanisms, and strict compliance with privacy regulations \citep{ryu2025vision,zhang2023challenges}.

Finally, specific architectures illustrate the breadth of foundational approaches. SAM \citep{kirillov2023segment} demonstrates strong zero-shot segmentation capabilities across modalities, CLIP \citep{radford2021learning} aligns visual and textual information for multimodal medical reasoning, and DINO \citep{caron2021emerging} provides self-supervised representations that adapt well to limited-label settings. In the language domain, models such as BERT \citep{kenton2019bert} and GPT \citep{brown2020language} contribute to medical report generation, text-based retrieval, and integration with imaging models. Together, these advances signal a paradigm shift: FMs are no longer just broad-purpose tools but are becoming key enablers of specialized clinical and surgical intelligence, as illustrated in Fig.~\ref{fig:foundational}.

\subsubsection{Segment Anything Model (SAM)}

The SAM \citep{kirillov2023segment} was developed to perform general-purpose image segmentation with minimal task-specific retraining using over 11 million images. Its adaptability extends to a wide array of applications, including autonomous driving, medical imaging, and satellite image analysis. Despite its non-medical origin, SAM shows considerable promise in healthcare settings. By combining robust CNN features and deep learning techniques, SAM can quickly adapt to different segmentation tasks, such as identifying tumors in CT scans, surgical tool recognition, or delineating organs in MRI data. This ability is referred to as zero-shot segmentation, as it allows SAM to perform inference on new imaging problems without extensive domain-specific training.

The subsequent iterations and specialized adaptations of SAM further illustrate its effectiveness and flexibility. For example, SAM3D adapts its architecture for 3D imaging tasks (e.g., CT, MRI, PET), using specialized 3D CNN layers \citep{bui2024sam3d}. In SAM, further innovations have been introduced that integrate multimodal imaging data, such as MRI and ultrasound, to achieve comprehensive diagnostic precision \citep{tan2024novelsam}. Further adaptations have been developed for the medical domain, such as the Medical SAM Adapter (MSA), which fine-tunes the SAM using adapter layers for higher segmentation precision in medical contexts \citep{chen2024ma}. SAM2 \citep{ravi2024sam}, a recent advanced version of SAM, has demonstrated significant improvements in performing real-time segmentation in videos, showcasing its enhanced capabilities in complex scenarios such as tumor identification and organ delineation \citep{ma2024benchmark}. Lastly, various efforts are being made to incorporate domain-specific knowledge into these models, enabling the capture of medical image nuances more effectively with limited data \citep{azad2023foundational}.

\begin{figure*}[hbtp]
    \centering
    \begin{subfigure}[b]{0.23\textwidth}
        \centering
        \includegraphics[width=\textwidth]{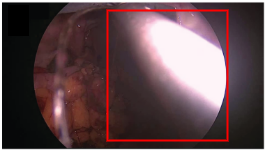}
        \caption{Instrument flare}
        \label{fig:sub1}
    \end{subfigure}
    \hfill
    \begin{subfigure}[b]{0.23\textwidth}
        \centering
        \includegraphics[width=\textwidth]{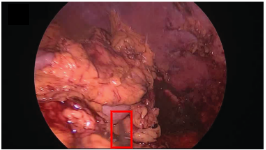}
        \caption{Partial occlusions by blood}
        \label{fig:sub2}
    \end{subfigure}
    \hfill
    \begin{subfigure}[b]{0.23\textwidth}
        \centering
        \includegraphics[width=\textwidth]{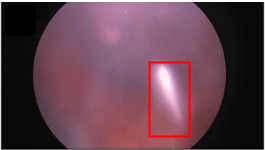}
        \caption{Smoke-induced occlusion}
        \label{fig:sub3}
    \end{subfigure}
    \hfill
    \begin{subfigure}[b]{0.23\textwidth}
        \centering
        \includegraphics[width=\textwidth]{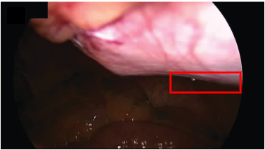}
        \caption{Underexposed regions}
        \label{fig:sub4}
    \end{subfigure}

    \begin{subfigure}[b]{0.23\textwidth}
        \centering
        \includegraphics[width=\textwidth]{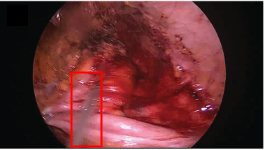}
        \caption{Motion blur}
        \label{fig:sub5}
    \end{subfigure}
    \hfill
    \begin{subfigure}[b]{0.23\textwidth}
        \centering
        \includegraphics[width=\textwidth]{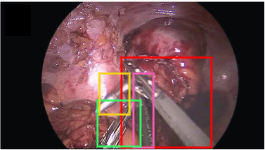}
        \caption{Multiple instruments}
        \label{fig:sub6}
    \end{subfigure}
    \hfill
    \begin{subfigure}[b]{0.23\textwidth}
        \centering
        \includegraphics[width=\textwidth]{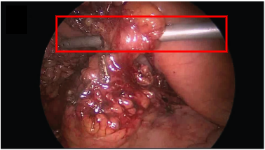}
        \caption{Partial occlusion by organ}
        \label{fig:sub7}
    \end{subfigure}
    \hfill
    \begin{subfigure}[b]{0.23\textwidth}
        \centering
        \includegraphics[width=\textwidth]{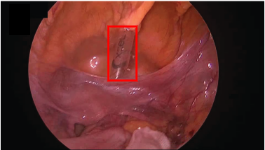}
        \caption{Transparent instrument}
        \label{fig:sub8}
    \end{subfigure}
\caption{\textit{Key challenges in surgical video analysis} include photometric artifacts (e.g., blurriness, specular reflections), occlusions from blood and tissue deformation, limited camera view, and tool similarity. These complexities hinder robust localization and segmentation, emphasizing the need for advanced Foundation Models. Image adapted from \citep{ceron2022real}.}
    \label{fig:chal}
\end{figure*}


\subsubsection{Contrastive Language-Image Pre-training (CLIP)}
CLIP \citep{radford2021learning} is a novel model introduced by OpenAI that revolutionizes the way machines understand images and text together. CLIP is trained in a contrastive manner where the images are paired with the corresponding text, learning from a wide variety of publicly available images and captions. This unique training approach allows CLIP to generalize across a wide array of visual concepts in a zero-shot manner. It can understand and categorize images that it has never seen before based on textual descriptions alone \citep{hafner2021clip}. In the medical and surgical fields, CLIP's capabilities can be particularly transformative. Its ability to interpret and correlate complex medical imagery with corresponding clinical notes or annotations without direct supervision makes it an excellent tool for diagnostic imaging. For example, CLIP can help radiologists and surgeons quickly identify relevant features in medical scans, such as magnetic resonance images or CT images, that correspond to textual descriptions found in case reports or diagnostic criteria. This could significantly speed up the diagnosis process, an imperative in the NHS FDS pathway, and improve the accuracy of identifying and classifying pathological characteristics. Moreover, CLIP's robust generalization ability allows it to adapt to diverse medical datasets, potentially reducing the time and resources required for model training and fine-tuning in specialized medical applications\citep{sun2023eva}.

\subsubsection{Self-Distillation with No Labels (DINO)}

DINO \citep{caron2021emerging} exemplifies a groundbreaking approach in self-supervised learning, initially designed for general computer vision tasks. Nevertheless, its ability to learn from unlabeled data makes it especially appealing for medical applications, where annotated data can be hard to obtain. By employing a teacher-student training framework, DINO captures meaningful image features from different augmented views of the same input without needing pre-existing annotations. This is very important for the medical domain as the data is often scarce and hard to annotate. 

In surgical contexts, for example, DINO could learn to recognize and segment important structures in endoscopic videos even if ground truth labels are scarce. This feature is crucial in clinical environments, where the time and expertise required to annotate large volumes of surgical footage may be limited. Many researchers have utilized DINO for different applications in several innovative ways. For instance, the SurgVID \citep{koksal2024surgivid} framework uses DINO's self-supervised learning to efficiently segment surgical tools and anatomical structures from video data, thus reducing the need for extensive manual annotations. This approach has proven to be nearly as effective as fully-supervised methods, showcasing the potential of self-supervised techniques in surgical applications.

Another adaptation, known as SurgicalDINO \citep{cui2024surgical}, modifies DINO for depth estimation in robotic surgery. This version incorporates Low-Rank Adaptation (LoRA) layers \citep{hu2021lora}, allowing the model to be fine-tuned for specific surgical applications without the need to retrain the entire model. This method demonstrates how DINO can be tailored to surgical navigation and 3D reconstruction. Additionally, \citep{ramesh2023dissecting} compared various self-supervised learning models, including DINO, on surgical datasets like Cholec80. They explored how well these models perform in recognizing surgical phases and detecting tools, indicating that self-supervised learning holds promise for enhancing surgical computer vision.

Furthermore, DINO's scalability and efficiency make it well-suited for medical centers that lack extensive computational resources. Through these advantages, DINO holds considerable promise in supporting tasks like surgical planning, intraoperative guidance, and postoperative analysis.

\subsubsection{Bidirectional Encoder Representations from Transformers (BERT)}
BERT is a pioneering language model that has significantly advanced natural language processing through its ability to understand context bidirectionally using a transformer-based architecture \citep{kenton2019bert}. In the surgical domain, the complexity and specificity of medical language require sophisticated models such as BERT, which are adept at capturing the nuanced semantics and intricate terminologies inherent in surgical texts. BERT's pre-training on extensive and diverse datasets enables it to develop a deep comprehension of language patterns, making it well-suited to address the rigorous demands of surgical data analysis \citep{bombieri2024surgicberta}. Its architecture facilitates an effective contextual representation, which is critical for accurately interpreting the multifaceted information present in surgical documentation and communication. Moreover, the adaptability of BERT through fine-tuning processes ensures that it can be tailored to meet the specific linguistic and contextual requirements of the surgical field, thereby enhancing the robustness and precision of language understanding in medical contexts. Although primarily focused on textual data, BERT also supports the integration and interpretation of findings from various medical imaging modalities, such as MRI or CT, by aiding in the automatic extraction and structuring of textual reports generated from these images, thus bridging the gap between radiological findings and surgical planning. This would be hugely useful in large-scale rapid automated reporting and would help alleviate the pressures human radiologists face when reporting large volumes of scans under the Faster Diagnostic Standards (FDS). In the UK, the FDS framework is a 28-day national health service (NHS) target to diagnose or exclude cancer in patients. These attributes position BERT as a Foundation Model in integrating advanced language processing capabilities within surgical data systems, contributing to the advancement of data-driven practices in surgical environments.

\subsubsection{Generative Pre-trained Transformer (GPT)}  
The GPT \citep{brown2020language} Foundation Model, developed by OpenAI, serves as a versatile backbone for numerous AI applications due to its exceptional generalization and natural language understanding capabilities. ChatGPT is the most prominent development by OpenAI, which is an instruction-tuned conversational model that adapts GPT's capabilities for interactive and user-centric applications. While GPT functions as a general-purpose Foundation Model capable of handling a wide range of language tasks, ChatGPT is fine-tuned to specialize in conversational contexts, making it particularly effective in medical settings \citep{van2025chatgpt,wang2024systematic}. It enables the automation of routine inquiries, enhances patient engagement through interactive dialogues, and streamlines clinical workflows by converting unstructured data into accurate and detailed medical reports. Moreover, its ability to assimilate and articulate complex medical literature makes it an invaluable tool for supporting research and continuing education for healthcare professionals \citep{xu2024current}. The integration of ChatGPT into clinical practice not only improves operational efficiency but also facilitates personalized patient care by providing timely and relevant medical information. However, deploying ChatGPT in healthcare requires rigorous attention to data privacy, ethical considerations, and model reliability to ensure secure and responsible use in sensitive medical environments.

Recent advances in GPT‐based large language models are rapidly extending into surgical applications, addressing both multimodal perception and domain‐specific knowledge tasks. Seenivasan et al. introduced SurgicalGPT \citep{seenivasan2023surgicalgpt}, an end‐to‐end trainable Language-Vision GPT (LV-GPT) that augments GPT-2 with a learnable vision tokenizer and token embeddings to perform visual question answering (VQA) in surgical scenes. \citep{cheng2023chatgpt} explored the broader impact of ChatGPT/GPT-4 \citep{achiam2023gpt} in surgical oncology and positioned these LLMs as versatile tools across the entire surgical workflow. They highlight potential applications in clinical trial design (e.g., feasibility analysis, sample‐size calculation), case management and data analysis (automated extraction of patient histories and outcome trends), and preoperative planning through AI-driven tumor delineation to reduce subjectivity in imaging interpretation. Furthermore, they envision GPT-4–powered support for multidisciplinary team consultations, virtual surgical simulations for trainee skill development, anatomical landmark navigation during operations, and even serving as the brain of future robot-assisted systems.

Furthermore,  \citep{beaulieu2024evaluating} provided a critical evaluation of the performance of GPT-4 \citep{achiam2023gpt} in standardized surgical knowledge assessments, comparing results in the Surgical Council on Resident Education (SCORE) bank and a second set of board-style questions referred to as (Data-B) containing 120 questions for practicing surgeons and senior surgical trainees. GPT-4 achieved human-level accuracy on multiple-choice items (71.3\% on SCORE; 67.9\% on Data-B) with high internal response concordance (88\%) and frequent non-obvious insights. However, its open-ended question performance was notably lower (47.9\% on SCORE), and repeat querying revealed answer variability in more than a third of initially incorrect responses, underscoring challenges in consistency and reliability for clinical deployment.



\subsubsection{Large Language Model Architecture (LLaMA)}
LLaMA \citep{touvron2023llama}, a high-capacity language model, leverages transformer-based mechanisms to understand and generate human language with remarkable accuracy and depth. Developed to facilitate advanced natural language processing tasks, LLaMA is designed for a wide array of applications ranging from automated text generation to complex query handling \citep{shao2024survey}. In the medical field, LLaMA's capabilities are particularly beneficial, as the model can interpret and synthesize medical literature, patient reports, and clinical guidelines with high precision. By training in various medical texts, LLaMA can help healthcare professionals by providing diagnostic suggestions, summarizing patient histories, and even generating informational content for patient education. Its ability to process and generate medical text effectively makes it an invaluable tool for enhancing clinical decision-making and improving patient outcomes. Moreover, LLaMA integration into clinical information systems can streamline workflows by automating documentation processes and extracting useful information from vast datasets, allowing medical personnel to focus more on patient care than administrative tasks \citep{nazi2024large}. As the demand for efficient and accurate processing of medical information grows, LLaMA stands out as a transformative model capable of revolutionizing various aspects of healthcare delivery.

\begin{table*}[ht]
\centering
\caption{Core surgical applications with standardized taxonomy, metrics, and limitations (reverse-chronological representative models).}
\label{tab:mlapplicationssurgery}
\resizebox{\textwidth}{!}{%
\begin{tabular}{@{}p{2.9cm}p{3.0cm}p{2.8cm}p{2.3cm}p{2.7cm}p{3.7cm}p{8.1cm}@{}}
\toprule
\textbf{Task} & \textbf{Subtask / Scope} & \textbf{Modality} & \textbf{Supervision} & \textbf{Method family} & \textbf{Representative model(s)} & \textbf{Metrics (typical)} \& \textbf{Common limitations} \\ \midrule

\textbf{Representation learning (foundation)} & Multi-modal pretraining from lecture videos & Endoscopic video + ASR transcripts & Self/unsup. (contrastive) & Dual-encoder VLM (CLIP-style) & SurgVLP~\citep{yuan2025learning} & Text–video retrieval (R@K), grounding (mAP), captioning (BLEU/CIDEr); strong zero-shot transfer to tools/phases, but ASR noise, text–video misalignment, and domain terminology limit fine-grained performance. \\

\textbf{Phase/Workflow recognition} & Online phase recognition in long procedures & Endoscopic video & Fully sup. & Transformers (LoViT: L-Trans + G-Informer) & LoViT~\citep{liu2025lovit} & Accuracy, F1 (e.g., +2.4 pp on Cholec80, +3.1 pp on AutoLaparo vs. prior SOTA); handles long-range context and phase transitions, but still compute-heavy and depends on high-quality phase labels. \\

\textbf{Tool tracking} & Multi-class multi-tool, multi-perspective tracking & Endoscopic video & Fully sup. (det+tracking) & Detector + re-ID + bipartite matching & SurgiTrack~\citep{nwoye2025surgitrack} & MOTA, IDF1, HOTA; real-time and robust re-ID via motion-direction/“operator” cues; challenges remain under heavy occlusion, out-of-body/out-of-view cycles, and look-alike tools; relies on CholecTrack20 annotations. \\

\textbf{Tracking} & Unsupervised segmentation under occlusion & Endoscopic video & Self-sup. & Neural (DL, clustering) & AMNCutter~\citep{sheng2025amncutter} & Dice, IoU; robust to occlusions, avoids pseudo-label noise. \\

\textbf{Segmentation (foundation/promptable)} & Pathology WSI (zero\text{-}shot) & Whole\text{-}slide images & Zero\text{-}shot / few\text{-}shot & Neural (DL, FM) & SAM (digital pathology)~\citep{deng2025segment} & Dice, IoU; large object masks OK, dense nuclei remain challenging; benefits from few\text{-}shot tuning. \\

\textbf{Segmentation} & Instrument masks (promptable/real-time) & Endoscopic video & Prompted / few\text{-}shot & Neural (DL, FM) & SurgSAM\text{-}2~\citep{liu2024surgical} & Dice, IoU; strong efficiency via frame pruning, but prompt sensitivity and domain shift remain. \\

\textbf{Event prediction / planning} & Target\text{-}conditioned action planning & Robotic surgical video & Supervised / diffusion & Neural (Diffusion) & See\text{-}Predict\text{-}Plan (MS\text{-}PCD)~\citep{zhao2024see} & Planning accuracy; improved foresight, but computationally heavy and needs precise goal specs. \\


\textbf{Segmentation/Depth} & Self-supervised depth \& segmentation & Robotic surgery video & Self-sup. & Neural (DL, FM) & SurgicalDINO~\citep{cui2024surgical} & Dice, Depth error; reduces annotation needs, but domain adaptation required. \\

\textbf{Segmentation} & General medical (foundation/promptable) & CT/MRI/US & Prompted / few-shot & Neural (DL, FM) & SAM~\citep{zhang2024segment}, U\text{-}Net variants~\citep{azad2024medical} & Dice, IoU; weak zero-shot in medical domains, tuning needed, prompt sensitivity. \\

\textbf{Tracking \& mapping} & Tool/scene tracking, SLAM & Endoscopy, colonoscopy & Self/weak sup. & Hybrid (SLAM + DL) & Schmidt et al.~\citep{schmidt2024tracking} & Precision, trajectory error; organ deformation, reflections, low texture, smoke. \\

\textbf{Workflow recognition (review)} & Phase/step analysis (survey) & Surgical video & N/A (review) & Survey (CNN/RNN/ViT) & Demir et al.~\citep{demir2023deep} & Accuracy, F1; highlights scaling issues and dataset bias; calls for standardization. \\

\textbf{Video understanding (representation)} & Self\text{-}supervision distillation & Surgical video & Self\text{-}sup. & Neural (distillation) & Free Lunch (Distilling Self\text{-}Supervisions)~\citep{ding2022free} & Accuracy, F1; reduces label needs, but distilled reps may miss fine, transient cues. \\



\textbf{Tool detection} & Bounding boxes / presence & Endoscopic video & Fully sup. & Neural (DL) & Bouget survey~\citep{bouget2017vision} & mAP, Precision/Recall; occlusions, motion blur, ambiguous small tools. \\

\textbf{Phase/Workflow recognition} & Step/phase sequence & Endoscopic video & Fully sup. & Neural (DL, temporal) & EndoNet/LSTM~\citep{twinanda2017endonet} & Accuracy, F1; long training times, error propagation across phases, weak temporal calibration. \\

\textbf{Anomaly/OoD detection} & Rare adverse events & Physio signals \& video & Self/unsup. & Neural (DL, AE) & Autoencoders~\citep{schlegl2017unsupervised} & AUC, AUROC, FPR95; high false positives, threshold instability, scarce positives. \\

\textbf{Skill assessment} & Dexterity/score prediction & Tool/motion tracks & Weak sup. & Classical ML & SVM (JIGSAWS)~\citep{ahmidi2017dataset} & Accuracy, MAE/RMSE; label subjectivity, small datasets, limited external validity. \\

\textbf{Event prediction} & Error/complication risk & OR operational data & Supervised & Classical ML & Random Forest~\citep{korbar2017deep} & Sensitivity, Specificity, AUC; overfitting, confounding variables, poor shift robustness. \\

\textbf{Segmentation} & Organ/structure masks & CT, MRI & Fully sup. & Neural (DL) & V\text{-}Net~\citep{milletari2016v} & Dice, IoU; large data needs, annotation burden, scanner/protocol variability. \\

\textbf{Depth estimation} & Dense depth from stereo & Endoscopic stereo & Fully sup. & Classical CV + Neural & Stereo vision~\citep{laina2016deeper} & Abs. rel. error, RMSE; calibration drift, specularities, low texture. \\

\textbf{3D recon. \& localization} & Scope/scene localization & Endoscopic video & Varied & Classical CV / Neural & Laparoscope localization~\citep{lin2016video} & Reproj. error, Precision; heavy compute, complex setup, limited real-time. \\

\textbf{Segmentation} & Instrument masks & Endoscopic video & Fully sup. & Neural (DL) & U\text{-}Net~\citep{ronneberger2015u} & Dice, IoU; class imbalance, specular highlights/occlusion, domain shift across centers. \\

\bottomrule
\end{tabular}
}
\end{table*}

\subsection{Challenges in Surgical Scene Understanding}
\label{sec:challenges}

Surgical scene understanding presents a diverse array of challenges that stem from the complexity and unpredictability of surgical environments, coupled with the technical limitations of current AI models. To advance AI-assisted surgical systems, the following challenges must be carefully addressed.

\subsubsection{Technical Challenges}

The domain of surgical video analysis presents unique challenges that hinder robust localization and segmentation of surgical instruments, as shown in Fig.~\ref{fig:chal}. Unlike natural images and videos, surgical frames are characterized by high tissue deformations and frequent occlusions caused by the presence of blood and multiple artifacts on the instruments. Photometric artifacts, as identified by \citep{ni2020pyramid}, can significantly degrade the performance of segmentation models. Some of the additional complexities include:
\begin{itemize}
 \item[a)] \textit{Subtle Variance and Limited View}: Surgical procedures may involve subtle interphase or intraphase variances that are difficult to capture consistently. The limited field of view offered by surgical visualization cameras further complicates visual assessment, while restricting the visual context available for decision-making \citep{hesamian2019deep}.
    \item[b)] \textit{Blurriness and Specular Reflection}:
Camera motion and the gas emissions from surgical tools often cause blurriness, while specular reflections and scale variations, as discussed by  \citep{baumhauer2008navigation}, can lead to poor segmentation accuracy.
\item[c)] \textit{Tool Similarity and Edge Presence}: The segmentation of multiple instruments is particularly challenging due to the appearance and shape similarity between different tools. Instruments located on the edge of video frames are especially difficult to detect and segment reliably \citep{karpat2008mechanics}. Moreover, the variations in the instrument pose may also alter the perceived geometry or shape, depending on the surgical camera's field of view.
\end{itemize}

\subsubsection{Lack of Representative Datasets}

Despite the proliferation of publicly available surgical video corpora \citep{li2024llava,jiang2025surgisr4k}, several concrete limitations undermine their utility for pretraining and evaluating large, generalist Foundation Models, which are discussed below.

\paragraph{Limited Scale and Procedural Diversity} Most of the datasets remain small and narrowly focused. For example, Cholec80 contains only 80 laparoscopic cholecystectomy videos from a single center \citep{twinanda2017endonet}, while JIGSAWS comprises 39 dry‑lab robotic suturing trials on phantom models \citep{gao2014jhu}. Even larger frame‑level collections such as the Dresden Surgical Anatomy Dataset (DSAD) cover primarily abdominal procedures \citep{kolbinger2023anatomy}. This lack of cross‑procedure breadth inhibits model generalization to rarer surgeries and atypical intra‑operative events.

\paragraph{Annotation Heterogeneity} Label schemas differ markedly across benchmarks: The M2CAI workflow defines coarse procedural phases \citep{jin2018tool}, JIGSAWS employs fine‑grained gesture taxonomies \citep{gao2014jhu}, and segmentation masks range from bounding boxes to pixel‑perfect expert contours \citep{grammatikopoulou2021cadis}. Very few datasets report inter‑annotator agreement or adhere to shared ontology standards, making it difficult to merge corpora or compare model performance across tasks.



\subsubsection{Ethical, Regulatory, and Privacy Challenges}

In addition to technical barriers, significant ethical, regulatory and privacy concerns must be addressed when applying AI to the understanding of the surgical scene. These considerations are essential to ensure that AI systems remain safe, reliable, and equitable in clinical settings. Below are some of the key challenges that require attention:

\begin{itemize}
    \item[a)] \textit{Patient Privacy and Data Protection:} The use of AI in surgical procedures frequently involves sensitive patient data, raising major concerns about privacy and data security. The protection of this information is vital to maintain trust among patients, clinicians, and institutions \citep{anderson2002ethics}.
    
    \item[b)] \textit{Regulatory Compliance and Bias:} AI systems in healthcare must comply with stringent regulatory standards, which continue to evolve \citep{char2018implementing}. Furthermore, AI models may introduce bias, especially if they are trained on nonrepresentative datasets, leading to disparities in care across different patient groups.
    
    \item[c)] \textit{Surgeon Accountability and Trust:} The introduction of AI into surgical decision making requires clear guidelines on surgeon accountability \citep{luxton2015artificial}. Although AI can assist in real-time decision making, surgeons must retain ultimate responsibility for patient outcomes. Ensuring that AI systems act as supportive tools rather than autonomous decision makers is critical to preserving trust and ethical responsibility.
    
\end{itemize}

\begin{takeaways}
Foundation Models (SAM/CLIP/DINO, LVLMs) are becoming the default substrate for surgical vision, but require careful domain adaptation due to privacy, compute, and workflow constraints. Multimodal fusion (video, audio, kinematics, text) is also emerging as the key to robust reasoning. Prompting/parameter-efficient tuning offers practical paths for limited data. Regulation and explainability remain gating factors for clinical use.
\end{takeaways}

\section{ML Applications in Surgical Tool Recognition}
\label{sec:tool_recognition}

The integration of machine learning (ML) technologies into surgical practices has catalyzed significant advances in computer-assisted interventions, improving both the precision and efficiency of procedures. These technologies facilitate a variety of applications, from diagnostic imaging to real-time operational guidance, fundamentally transforming the surgical landscape. The systematic deployment of ML methodologies has great potential in improving surgical outcomes as well as streamlining both preoperative planning and postoperative care. 

As detailed in Table~\ref{tab:mlapplicationssurgery}, ML applications in surgery use a diverse range of techniques, including supervised learning for tissue classification, unsupervised anomaly detection, and deep learning (DL) architectures for more complex tasks such as instrument segmentation and surgical skill evaluation. In this manuscript, we treat DL as a subset of ML, and therefore collectively refer to all methods under the umbrella of ML unless the distinction is explicitly relevant. 

These techniques employ various models such as convolutional neural networks (CNNs) for image-based analysis, segment-anything models (SAMs) for zero-shot segmentation, and recurrent neural networks (RNNs) for time-series data, addressing specific needs within the surgical workflow. Table~\ref{tab:mlapplicationssurgery} categorizes these applications by specifying the ML methods used with the corresponding models and the data types utilized, ranging from histopathological images to surgical video data. It also outlines the metrics used to evaluate performance, such as accuracy, precision, recall, and the intersection over union (IoU). In particular, the table acknowledges the limitations inherent to each application, such as the high computational costs and the need for extensive data annotation, which are critical considerations for future research directions.

\subsection{Instrument Segmentation}
Instrument segmentation is particularly crucial in robot-assisted surgery (RAS) because the surgeon operates through a console and depends entirely on the endoscopic video feed for spatial awareness. Unlike conventional surgery, there is no direct view of the operative field, making precise localization of surgical instruments essential. Accurate segmentation enables downstream tasks such as tool tracking, motion analysis, and collision avoidance, and it supports augmented-reality overlays that link intraoperative video with preoperative imaging. These capabilities directly affect surgical precision, reduce the risk of inadvertent tissue injury, and maintain the overall safety and efficiency of RAS workflows.

A significant contribution to this field is detailed by García-Peraza-Herrera et al. \citep{garcia2017real}, who explore CNN-based methods for real-time surgical tool segmentation in laparoscopic videos. This study underscores the improvements deep learning (DL) models bring to surgical tool segmentation, improving clarity and operational efficiency in robotic surgeries, which are critical to ensuring safety and precision.

In Table~\ref{table:instrumentsegmentation}, the advanced models are listed for instrument segmentation, covering a spectrum from supervised learning with models like FCN and U-Net to complex methods involving adversarial networks and domain adaptation. These models are evaluated on datasets such as EndoVis and various private datasets, utilizing techniques that combine CNNs with RNNs and residual networks. Transitioning from traditional segmentation techniques, the Min-max Similarity model introduces a novel contrastive and semi-supervised learning framework that marks a significant advancement in medical imaging. This model addresses the challenge of limited annotated data \citep{lou2023min} by maximizing the similarity between representations of similar objects while minimizing those of dissimilar ones. The innovative approach improves the ability to distinguish between surgical tools and complex backgrounds, leveraging unlabeled data effectively to overcome the scarcity of labeled datasets. Its semi-supervised nature facilitates generalization across different surgical tools and procedures, reducing the need for extensive retraining. This method not only advances technical capabilities but also provides substantial clinical benefits by enhancing the autonomy of robotic surgical systems and potentially reducing the cognitive load on surgeons, thus increasing surgical safety.

However, label efficiency continued to be the central bottleneck in 2025, and two complementary directions stood out. SegMatch \citep{wei2025segmatch} reframes FixMatch-style \citep{sohn2020fixmatch} semi-supervision for endoscopic images, pairing consistency regularization with pseudo-labeling tailored to surgical textures and illumination. It consistently boosts Dice/IoU under limited labels, and shows better cross-domain generalization than fully supervised baselines, making a strong case for semi-supervision in minimally invasive surgery (MIS) segmentation. In parallel, AMNCutter \citep{sheng2025amncutter} demonstrates label-free (unsupervised) instrument segmentation via an affinity and attention-guided multi-view normalized-cut objective, avoiding error-prone pseudo-masks and proving robust in smoky or occluded scenes, which are often observed in surgical videos. Together, these contributions point to a pragmatic future with semi- or unsupervised pre-training to shrink annotation budgets, followed by small amounts of task- or site-specific supervision. Another application is to use perception to unlock geometry without onerous calibration. WS-SfMLearner \citep{lou2024ws} removes the long-standing assumption of known camera intrinsics by jointly learning depth, ego motion, and intrinsics from monocular surgical videos with self-supervision of the cost volume. 

Building on these advances, the development of Efficient Class-Promptable Surgical Instrument Segmentation represents a further evolution in the field. Models like the SAM prompting system can be directed toward segment-specific classes of surgical instruments through simple prompts or minimal user input, such as drawing a bounding box or a point prompt of the target object \citep{yue2024surgicalsam}. This capability is crucial in surgical settings, where the types of tools and the visual environment can vary significantly between procedures. Such models allow for rapid adaptation to new instruments without the need for extensive retraining, improving their practical utility in the operating room. This adaptability not only increases segmentation accuracy but also reduces computational demands, making real-time implementation more feasible \citep{twinanda2016endonet}. 

Further enriching the field, advances in Foundation Models, particularly those based on Transformer architectures, have significantly enhanced surgical image segmentation. These models utilize self-attention mechanisms to effectively manage complex spatial relationships in medical imaging, complementing traditional techniques like CNNs \citep{dosovitskiy2021imageworth16x16words, dosovitskiy2020image}. Key innovations, such as SAM, demonstrate the adaptability required for dynamic surgical environments, crucial for applications where rapid changes demand reliable model performance \citep{kirillov2023segment, zhou2019fast}. Moreover, the integration of these advanced models into augmented reality systems provides surgeons with context-sensitive real-time information that improves both precision and safety during surgical procedures \citep{kunz2022augmented}. These would act as clinical decision support tools that highlight features linking real-time images with preoperative CT or MRI scans. For example, pre-operative CT detection of an enlarged lymph node with potential cancer cells in the mesentery next to the ileo-colic or middle colic artery could be highlighted in real time when performing an extended right hemicolectomy (removal of the right and transverse colon). This would help the surgeon achieve a clear surgical margin and reduce the risk of cancer recurrence. 

The recent development of the Video-Instrument Synergistic Network (VISN) underscores a significant advancement in leveraging machine learning for robotic surgeries, as explained by Wang et al. \citep{wang2024video}. VISN is specifically designed to handle the complexities of video-based instrument segmentation in this field by automatically identifying and segmenting surgical instruments based on natural language expressions. By exploiting the synergistic interactions between video frames and instrument-specific features, it significantly enhances segmentation accuracy. Beyond accuracy, VISN contributes to the safety and efficacy of RAS by maintaining temporal consistency across frames, ensuring instruments remain clearly identified despite occlusions, smoke, or rapid motion. This reduces the likelihood of visual misinterpretation and relieves the surgeon from mentally reconstructing instrument trajectories across frames. By automating routine perception tasks such as tool presence detection and motion tracking, VISN directly lowers cognitive load, allowing surgeons to concentrate on higher-level decision-making. This integration of stable perception into surgical workflows thus enhances both precision and patient safety in complex robotic procedures.

\begin{table}[hbtp]
  \centering
  \caption{Segmentation models for surgical tools and instruments within various surgical environments, as seen in robotic and laparoscopic surgeries, sorted from 2025 onwards.}
  \label{table:instrumentsegmentation}

  \resizebox{0.9\columnwidth}{!}{%
  \begin{tabular}{>{\raggedright\arraybackslash}p{0.35\columnwidth} >{\raggedright\arraybackslash}p{0.35\columnwidth} >{\raggedright\arraybackslash}p{0.20\columnwidth} >{\raggedright\arraybackslash}p{0.20\columnwidth}}
    \hline
    \textbf{Reference} & \textbf{Dataset} & \textbf{Model} & \textbf{Technique} \\
    \hline
    \citep{wei2025segmatch} & EndoVis datasets & SegMatch & Semi-Supervised Consistency \\
    \citep{sheng2025amncutter} & EndoVis datasets & AMNCutter & Unsupervised Segmentation \\
    \citep{rahman2024emcad} & EndoVis datasets & EMCAD & Attention-enhanced Decoder \\
    \citep{peng2024reducing} & UW Sinus, EndoVis17 & DeepLabv3+, MobileNet & Active-Learning \\
    \citep{lou2023min} & EndoVis datasets & Min-max Similarity Model & Semi-Supervised \\
    \citep{zhao2021anchor} & Davis16, EndoVis17,18, HKPWH & AOMA & Meta-Learning \\
    \citep{su2021local} & Sinus Surgery & CycleGAN & Adversarial \\
    \citep{zhang2021surgical} & Private, EndoVis17 Synthetic & U-Net+PatchGAN & Adversarial \\
    \citep{sahu2020endo} & Sim, Sim Cholec80, EndoVis15 & DNN+, Ternaus11 & Unsupervised Domain Adaptation \\
    \citep{gonzalez2020isinet} & EndoVis17,18 & ISINet & Supervised \\
    \citep{ni2020barnet} & Cata7, EndoVis17 & BarNet & Supervised \\
    \citep{colleoni2020synthetic} & Custom & FCNN & Supervised \\
    \citep{lin2020lc} & Sinus Surgery & LC-GAN & Adversarial \\
    \citep{yu2020holistically} & EndoVis17 & Modified U-Net & Supervised \\
    \citep{jin2019incorporating} & EndoVis17 & MF-TAPNet & Supervised, Self-supervised \\
    \citep{mohammed2019streoscennet} & EndoVis17 & StreoScenNet & Supervised \\
    \citep{islam2019real} & EndoVis17 & CNN+Residual & Auxilary, Adversarial \\
    \citep{lee2019weakly} & Private & DCNN & Weakly-Supervised \\
    \citep{fuentes2019easylabels} & EndoVis15 & DeepLabv3+ & Weakly-Supervised \\
    \citep{pakhomov2019deep} & EndoVis15 & ResNet+atrous & Supervised \\
    \citep{ross2018exploiting} & EndoVis17 & ResNet, U-Net & Semi-Supervised \\
    \citep{milletari2018cfcm} & EndoVis15 & ResNet +, Conv LSTM & Supervised \\
    \citep{shvets2018automatic} & EndoVis-17 & Ternaus11, Ternaus16, LinkNet34 & Supervised \\
    \citep{garcia2017toolnet} & DVR & ToolNet & Supervised \\
    \citep{garcia2017real} & EndoVis15, NST, FFT & FCN-8s+ & Supervised \\
    \citep{attia2017surgical} & EndoVis15 & CNN + RNN & Supervised \\
    \hline
  \end{tabular}%
  }
\end{table}

\begin{table}[h!]
  \centering
  \caption{Comprehensive overview of models for instrument detection for surgical tools in dynamic surgical settings, showcasing their application on different datasets.}
  \label{table:instrumentdetection}
  \resizebox{\columnwidth}{!}{%
  \begin{tabular}{>{\raggedright\arraybackslash}p{0.30\columnwidth} >{\raggedright\arraybackslash}p{0.20\columnwidth} >{\raggedright\arraybackslash}p{0.35\columnwidth} >{\raggedright\arraybackslash}p{0.30\columnwidth}}
    \hline
    \textbf{Reference} & \textbf{Dataset} & \textbf{Model} & \textbf{Technique} \\
    \hline
    \citep{nwoye2025surgitrack} & CholecTrack20, EndoVis & SurgiTrack & Detector + re-ID + bipartite matching (tracking/detection) \\
    \citep{teevno2023semi} & m2cai16 & Teacher-Student & Semi-Supervised \\
    \citep{zhang2021surgical} & AJU-Set, m2cai16-tool-locations & FasterRCNN+, Region proposal Network & Supervised \\
    \citep{kondo2021lapformer} & Cholec80 & CNN + Transformer & Supervised \\
    \citep{alshirbaji2021deep} & Cholec80 & CNN+LSTM & Supervised \\
    \citep{yoon2020semi} & Private & Faster, CascadeRCNN & Semi-supervised \\
    \citep{namazi2022contextual} & M2CAI16, Cholec-80 & RCNN & Supervised \\
    \citep{vardazaryan2018weakly} & Cholec80 & FCN & Weakly-supervised \\
    \citep{hu2017agnet} & m2cai16-tool & AGNet & Supervised \\
    \citep{mishra2017learning} & m2cai16-tool & CNN+LSTM & Supervised \\
    \citep{kurmann2017simultaneous} & RMIT, EndoVis15 & CNN & Supervised \\
    \citep{sarikaya2017detection} & ATLASDione & CNN + RPN & Supervised \\
    \citep{twinanda2016endonet} & Cholec80, EndoVis15 & EndoNeT & Supervised \\
    \hline
  \end{tabular}

  }
\end{table}

\subsection{Instrument Detection}
Instrument detection in surgical settings is crucial for identifying surgical tools' presence and location within images or video frames, setting the foundational for tracking movements and recognizing surgical activities. Most of the detection models leverage sophisticated frameworks like Faster R-CNN or YOLO for efficient multi-object detection in the dynamic environments of surgical scenes.

\citep{twinanda2016endonet} introduced the EndoNet architecture, combining tool detection with phase recognition, exemplifying how multitask learning enhances surgical scene understanding by simultaneously tackling related tasks, thus improving the robustness and accuracy of tool identification and contextual understanding. In open surgery, DL models detect and localize tools within dynamic environments, which helps postoperative analysis and real-time decision making \citep{fujii2022surgical}. This capability not only improves surgical safety and efficiency by ensuring the correct use and placement of instruments, but also enhances surgical training by providing real-time feedback on tool handling.

Furthermore, integrating instrument detection with phase recognition, as further developed by \cite{twinanda2017endonet}, demonstrates the potential of combining multiple analytical tasks for an understanding of surgical workflow. Such systems contribute to automated documentation, improved situational awareness, and improved coordination of surgical procedures. Table \ref{table:instrumentdetection} summarizes various state-of-the-art models for instrument detection, including techniques and models like EndoNet, Faster R-CNN, and Cascade R-CNN in datasets such as Cholec80, EndoVis15, RMIT, and m2cai16-tool. This table illustrates the evolution of instrument detection technology, emphasizing advancements towards sophisticated, multitask learning frameworks that promise greater precision in real-time surgical tool detection.

In addition to these applications, instrument detection technologies are instrumental in advancing robot-assisted surgeries. Accurate detection and localization of surgical tools enable precise control and manipulation by robotic systems, thereby enhancing the surgeon's capabilities and reducing the likelihood of errors. The ability to monitor tool positions in real-time also facilitates the implementation of safety protocols, such as preventing unintended movements and ensuring that instruments are used within their designated areas.

\subsection{Instrument Tracking}
Instrument tracking in surgical environments involves continuous monitoring of surgical tools across video frames, critical for the effective operation of robotic systems and assistive technologies. This capability ensures accurate real-time localization of instruments, adapting seamlessly to surgeon actions and enhancing procedural safety and efficiency.

Table \ref{table:instrumenttracking} provides an overview of advanced state-of-the-art models for instrument tracking, including LinkNet and ST-MTL, which excel in complex tasks combining detection, segmentation, and tracking, essential for precise real-time instrument tracking in modern surgeries. Other models like U-Net and FCN are noted for their robust detection and accurate instrument positioning, which are crucial for maintaining performance in dynamic surgical settings. The table details the effectiveness of these models across diverse surgical environments, listing the specific datasets validated along with the tracking techniques used. This comprehensive information highlights how each model contributes to instrument tracking, offering insights into their operational efficiencies and applications.

\begin{table}[hbtp]
  \centering
  \caption{Tracking methods for surgical instruments across video frames, highlighting their effectiveness in dynamic surgical environments.}
  \label{table:instrumenttracking}
  \resizebox{0.98\columnwidth}{!}{%
  \begin{tabular}{>{\raggedright\arraybackslash}p{0.30\columnwidth} >{\raggedright\arraybackslash}p{0.20\columnwidth} >{\raggedright\arraybackslash}p{0.25\columnwidth} >{\raggedright\arraybackslash}p{0.40\columnwidth}}
    \hline
    \textbf{Reference} & \textbf{Dataset} & \textbf{Model} & \textbf{Technique} \\
    \hline
    \citep{sheng2025amncutter} & EndoVis datasets & AMNCutter & Unsupervised segmentation-based tracking under occlusion \\
    \citep{teevno2023semi} & m2cai16 & Teacher-Student & Semi-supervised detection + tracking consistency \\
    \citep{islam2021st} & EndoVis17 & ST-MTL & Tracking, Segmentation \\
    \citep{zhao2019real} & Private & CNN & Detection \\
    \citep{nwoye2019weakly} & Cholec80 & FCN + ConvLSTM & Detection \\
    \citep{lejeune2018iterative} & BRATS, EndoVis15, Cochlea & U-Net & Features \\
    \citep{du2018articulated} & RMIT, EndoVis15 & FCN & Detection - Regression \\
    \citep{sarikaya2017detection} & ATLASDione & CNN+RPN, Fast RCNN & Detection \\
    \citep{zhang2017real} & m2cai16-tool & LinkNet & Detection and Segmentation \\
    \citep{allan2015image} & Self & Decision tree, Optical flow & Features \\
    \hline
  \end{tabular}

}
\end{table}

\paragraph{Traditional Tracking Methods}
Traditional tracking methods such as the Kalman Filter are widely used in dynamic systems tracking due to their ability to predict state variable changes based on velocity and acceleration measurements. This method excels in environments with linear motion profiles and Gaussian noise, which can be reliably estimated under these specific conditions \citep{kalman1960new}. It uses a predictive model that iteratively updates state estimates, making it suitable for applications such as surgical tool tracking, where motions are typically linear. However, surgical instruments often display non-linear movement patterns with sudden changes in direction and speed, challenging traditional tracking methods. The assumptions of linearity and Gaussian noise integral to Kalman Filters may not apply in surgical contexts, leading to potential inaccuracies and impairing the filter's effectiveness in complex surgical maneuvers \citep{dissanayake2001solution}.



\paragraph{DL-based Methods}
To address the limitations of traditional tracking methods, advanced DL-based methodologies using neural networks have been developed, utilizing temporal consistencies and contextual data to improve the accuracy of the tracking. DL architectures like RNNs and Long Short-Term Memory (LSTM) networks excel in capturing complex temporal dependencies and adapting to the unpredictable dynamics of surgical tool movements \citep{liu2023reproducing}. These models trained on extensive datasets of annotated surgical activities, generalize well across different surgical scenarios and instrument types. A key development by Allan et al. \citep{iovene2024hybrid} uses DL for real-time surgical tool tracking, combining CNNs with LSTM layers to effectively model the spatial and temporal aspects of surgical instruments. This hybrid approach adeptly manages rapid movements and visual occlusions, which are common challenges in surgical settings, enhancing both the functionality of surgical assistive technologies and the precision of robotic instrument control, thus reducing operational errors and improving outcomes.

Furthermore, recent studies have explored Transformer-based models, which are renowned for handling long-range dependencies within sequential data \citep{vaswani2017attention}. These models, with their scalability and adaptability, are well-suited for the complex data involved in surgical instrument tracking. Transformers employ attention mechanisms to selectively focus on relevant data segments, significantly enhancing tracking accuracy across extended periods and diverse surgical environments.

\paragraph{Sensor-Based and Integrated Tracking Systems}
Sensor-based tracking systems augment vision-centric methodologies by incorporating data from electromagnetic trackers or Inertial Measurement Units (IMUs), adding dimensions of information that improve the robustness of the tracking \citep{nakawala2017toward}. These multimodal approaches counteract the limitations of vision-based systems, such as occlusions and variable lighting, by providing additional reliable sources of positional data. Integration of instrument tracking with detection and segmentation technologies is also critical, as it facilitates consistent and accurate representations of the surgical environment and allows functionalities such as automated tool re-identification, workflow analysis, and Augmented Reality (AR) overlays for surgical navigation \citep{ayobi2024pixel}. Despite advances, tracking in dynamic and cluttered surgical settings remains challenging due to factors such as tool occlusions, variable lighting, and the presence of biological tissues, along with the need for real-time processing, which imposes directions on computational efficiency. Therefore, developing sophisticated models that balance tracking accuracy with computational speed is essential to maintain robust performance without disrupting surgical workflows.

\begin{figure*}[hbtp]
    \centering
    \begin{subfigure}[b]{0.45\textwidth}
        \centering
        \includegraphics[width=\textwidth]{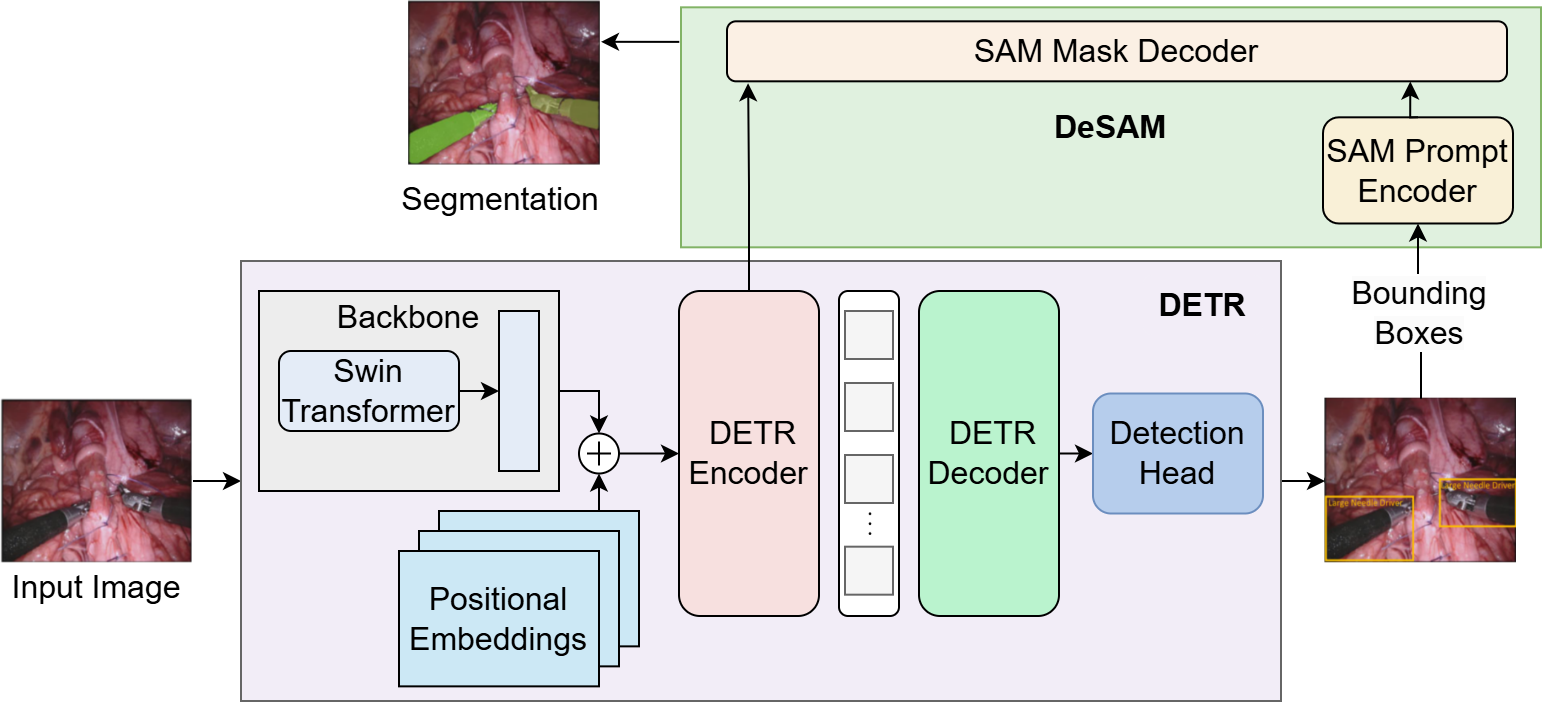}
        \caption{DeSAM \citep{sheng2024surgical}}
        \label{figSAM:sub1}
    \end{subfigure}
    \hfill
    \begin{subfigure}[b]{0.45\textwidth}
        \centering
        \includegraphics[width=\textwidth]{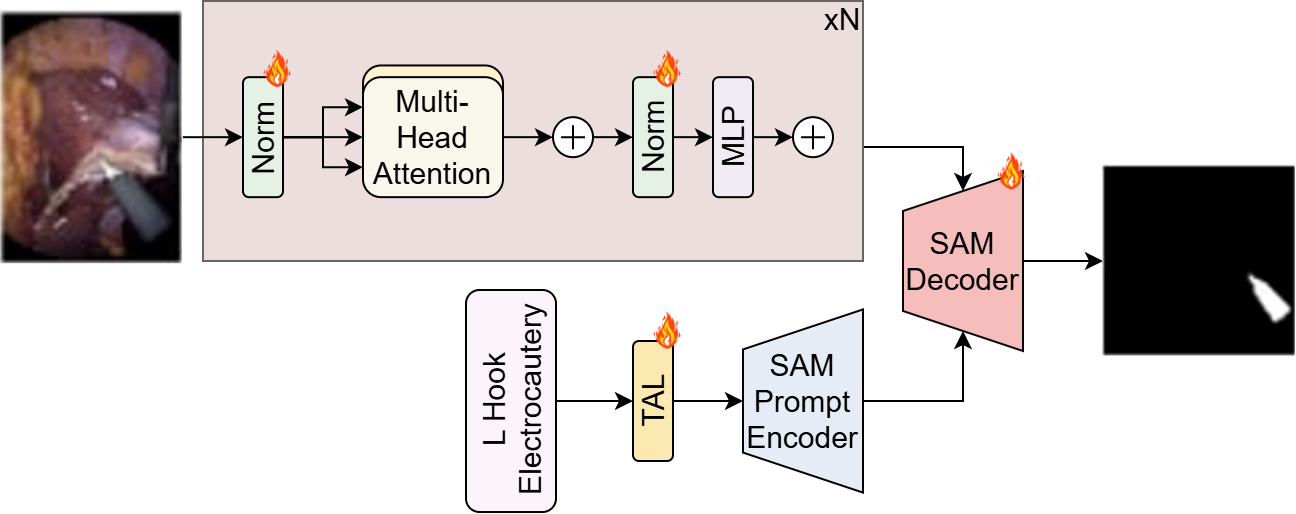}
        \caption{AdaptiveSAM \citep{paranjape2024adaptivesam}}
        \label{figSAM:sub2}
    \end{subfigure}
    
    \begin{subfigure}[b]{0.45\textwidth}
        \centering
        \includegraphics[width=\textwidth]{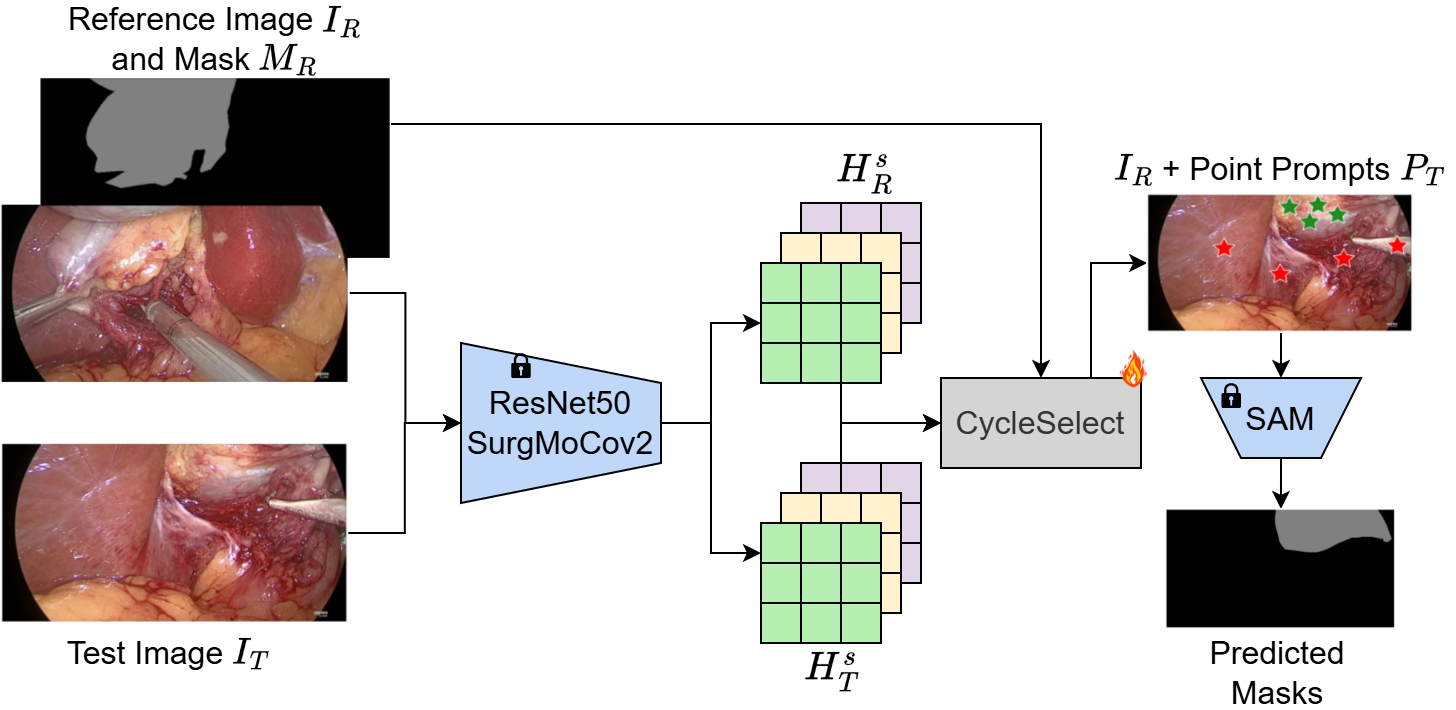}
        \caption{CycleSAM \citep{murali2024cyclesam}}
        \label{figSAM:sub3}
    \end{subfigure}
    \hfill
    \begin{subfigure}[b]{0.45\textwidth}
        \centering
        \includegraphics[width=\textwidth]{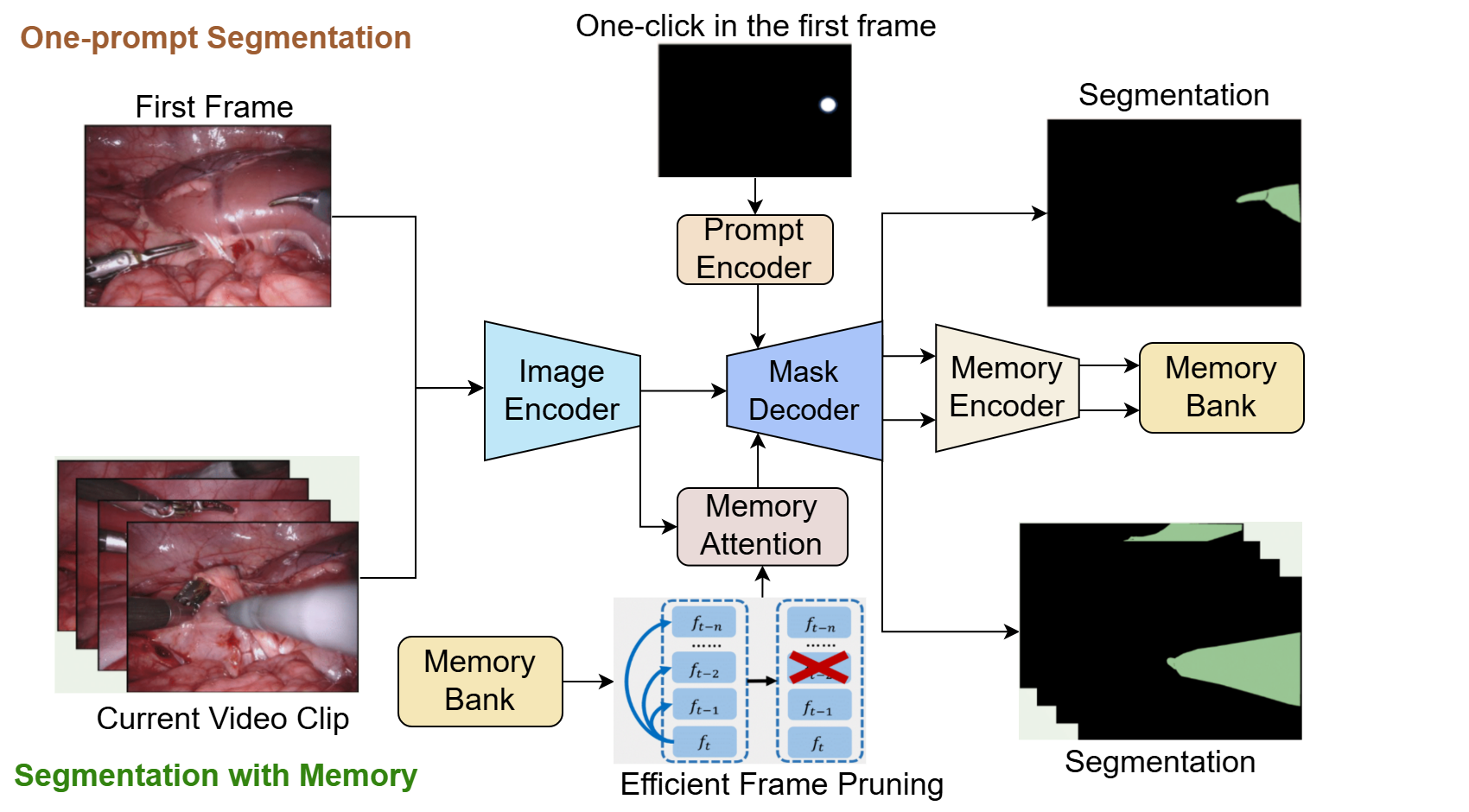}
        \caption{Surgical SAM2 \citep{liu2024surgical}}
        \label{figSAM:sub4}
    \end{subfigure}
\caption{Segmentation tasks in surgical video analytics addressed by SAM variants: (a) precise tool localization via detection and segmentation integration (\textit{DeSAM}); (b) improved segmentation in complex scenarios with task-specific prompts and multi-head attention (\textit{AdaptiveSAM}); (c) temporal consistency through reference-based segmentation (\textit{CycleSAM}); and (d) efficient video processing with memory-based frame pruning (\textit{Surgical SAM2}). These adaptations highlight SAM's versatility in tackling the challenges of surgical environments. Images adapted from corresponding references.}    
\label{fig:SAM}
\end{figure*}

\subsection{Instrument Pose Estimation}


Pose estimation in surgical settings involves traditional computer vision and modern deep learning-based methods. Traditional techniques using feature detection and geometric modeling often employ regression-based approaches such as Support Vector Regression (SVR) and Random Forests to predict pose parameters directly from images \citep{kehl2017real}. However, these methods may struggle with the variability of the surgical environment, such as instrument occlusions and various orientations. Deep learning advances have greatly improved the accuracy and robustness of pose estimation. Fully Convolutional Networks (FCNs) are used for articulated pose estimation, detecting multiple key points and their spatial relationships \citep{du2018articulated}. Furthermore, RNNs and Long Short-Term Memory (LSTM) networks also incorporate temporal information from video sequences to enhance real-time pose tracking \citep{laina2017concurrent}. These architectures manage temporal dependencies and motion dynamics, ensuring consistent pose estimations across frames, even in the presence of motion blur and occlusions.


Accurate instrument pose estimation is critical to improving various aspects of surgical practice. In robot-assisted surgeries, it provides precise control of robotic arms, allowing surgeons to execute complex maneuvers with a reduced risk of instrument clash \citep{cepolina2022introductory}. Pose estimation also supports AR systems in operating rooms by overlaying essential information onto the surgeon's field of view, thus improving situational awareness and navigation \citep{chidambaram2021applications}. Furthermore, in surgical training and simulation, it enables the creation of realistic simulators that offer real-time feedback on instrument handling and techniques, improving training effectiveness \citep{zhao2019real}. Despite its benefits, pose estimation in surgical settings faces challenges such as tool occlusions, variable lighting, and complex backgrounds. Real-time performance without compromising accuracy remains essential for practical application. Future research aims to enhance the robustness and adaptability of pose estimation systems through the use of unsupervised and semi-supervised learning techniques, the integration of multimodal data, and the incorporation of domain-specific knowledge to meet the unique requirements of surgical environments \citep{nakawala2017toward}.



\subsection{Anomaly Detection and Safety Measures}
Anomaly detection and safety measures use ML to identify deviations from normal operations that can indicate hazardous situations, in order to enhance surgical safety through alerts or corrective actions. \citep{kayan2024casper} explored how the integration of anomaly detection systems in robotic surgery can preemptively address risks by recognizing unusual tool movements or unsafe interactions.

\subsubsection{Complication Detection}

Complication Detection focuses on identifying potential complications during surgeries, such as unexpected bleeding or improper instrument handling, using machine learning to analyze real-time surgical data.\citep{ruan2022real} highlight continuous monitoring of the surgical process, integrating data from surgical tools and physiological monitors to enhance the detection capabilities of complications.
\subsubsection{Error Prediction} 

Error Prediction forecasts potential errors in the handling or functioning of surgical instruments to enhance safety and efficiency.\citep{miao2024predictive} demonstrate how DL models can predict errors during robotic surgeries in real time, effectively reducing error rates by alerting surgeons to potential instrument misuse or failure before they compromise the procedure.


\subsection{Segmentation using Foundation Models}

Building upon the foundational laid by traditional ML and DL techniques in surgical applications, this section shifts focus to Foundation Models, specifically the Segment Anything Model (SAM). In surgical applications, SAM is being refined for precise tool segmentation. This involves fine-tuning the model with surgical-specific datasets and incorporating domain-specific knowledge to enhance precision without sacrificing generalization across different surgical setups \citep{oguine2024generalization}. High-resolution surgical images and detailed annotations are used in training to help the model learn the nuanced features of surgical instruments and tissues \citep{shvets2018automatic}. Techniques such as transfer learning and domain adaptation adjust the model's parameters for surgical contexts, with few-shot learning enabling performance improvements with limited data \citep{twinanda2016endonet}. These advancements contribute to more precise segmentation outputs, which is crucial for real-time instrument tracking, surgical navigation, and augmented reality applications in the operating room \citep{deng2023segment}. Below are some variants of the SAM architecture as shown in Fig \ref{fig:SAM}:

\subsubsection{Surgical-DeSAM}
The Surgical-DeSAM explores the decoupling of the SAM's components to better suit robotic surgical environments, where precision and reliability are paramount. By modularizing the segmentation tasks, DeSAM enhances the robustness and accuracy of instrument segmentation, which is critical for automated or semi-automated robotic surgeries \citep{sheng2024surgical}. Decoupling allows each module to be fine-tuned individually, enabling more precise control over the segmentation process and facilitating the integration of domain-specific knowledge relevant to robotic surgery \citep{qu2021surgical}. In such robotic surgeries, accurate instrument segmentation is essential for tasks such as motion tracking, collision avoidance, and providing visual feedback to the surgeon \citep{lin2020lc}. The complexity of surgical scenes, combined with the variability of instruments and tissues, presents significant challenges for segmentation models. By decoupling SAM's components, Surgical-DeSAM addresses these challenges by allowing for specialized processing of different aspects of the segmentation task. For instance, separate modules can handle the unique visual features of robotic instruments or adapt to varying lighting conditions in the surgical environment \citep{ross2018exploiting}. This modular approach also facilitates easier updates and maintenance of the system, as individual components can be improved or replaced without affecting the entire model \citep{eppler2023automated}.

\subsubsection{AdaptiveSAM}
AdaptiveSAM \citep{paranjape2024adaptivesam} represents an evolution in the application of the SAM for surgical segmentation, focusing on dynamically adjusting the segmentation parameters in response to the specific needs of each surgical video frame. This adaptability ensures high precision and relevance of the segmentation output, adapted to the specific characteristics of the surgical regions \citep{paranjape2024adaptivesam}. In surgical environments, conditions such as lighting, tissue appearance, and instrument presence can change rapidly, making static segmentation models less effective \citep{twinanda2017endonet}. AdaptiveSAM addresses this challenge by incorporating mechanisms that fine-tune the model on the fly, enhancing its responsiveness to the dynamic surgical scene. One of the key features of AdaptiveSAM is its ability to efficiently utilize computational resources while maintaining high segmentation accuracy \citep{yang2022tmf}. By employing techniques like transfer learning and incremental learning, the model can adapt to new data without retraining from scratch \citep{ouyang2022self}. This is particularly important in surgical settings where real-time performance is crucial and computational resources may be limited. AdaptiveSAM's efficient tuning process allows it to provide immediate feedback to surgeons, assisting in tasks such as instrument tracking, tissue identification, and navigation .

\subsubsection{CycleSAM}
CycleSAM introduces a novel approach that leverages cycle-consistent feature matching to enable one-shot learning within the SAM, allowing the model to perform accurate segmentation from a single annotated example. This capability is particularly valuable in surgical settings where obtaining comprehensive training data is challenging due to the scarcity of annotated images and the high cost of expert labeling \citep{murali2024cyclesam}. In traditional segmentation models, a large amount of annotated data is required to achieve high accuracy, which is often impractical in medical contexts. CycleSAM addresses this limitation by using cycle-consistent adversarial networks to align features between the source domain (annotated examples) and the target domain (unlabeled surgical images) \citep{zhu2017unpaired}. This method ensures that the learned features are robust and transferable, enabling the model to generalize from a single example to a variety of surgical scenes. The loss of cycle consistency enforces a bidirectional mapping between the domains, preserving the structural integrity of surgical images while adapting to new contexts \citep{hoffman2018cycada}. This approach effectively mitigates the domain shift problem commonly encountered in medical image analysis, where variations in imaging conditions can significantly impact model performance \citep{chen2019synergistic}. By incorporating cycle-consistent feature matching, CycleSAM enhances SAM's ability to perform accurate segmentation without extensive retraining or additional data collection. This one-shot learning capability is crucial in surgical environments, where real-time decision-making is essential and the availability of annotated data is limited. CycleSAM's efficiency enables quicker deployment of segmentation models, facilitating applications such as surgical navigation, instrument tracking, and intraoperative guidance \citep{maier2017surgical}.  

\subsubsection{Surgical SAM 2}
Advancing the capabilities of the SAM, recent iterations focus on the real-time segmentation of surgical videos by employing efficient frame pruning techniques. These innovations reduce the computational load by selectively processing frames that contain significant changes or relevant surgical actions, enabling SAM to operate effectively in real-time surgical scenarios \citep{liu2024surgical}. This is crucial in dynamic surgical environments where rapid processing is essential for assisting surgeons without disrupting the workflow. By minimizing latency and ensuring timely feedback, efficient frame pruning enhances the practicality of SAM in the fast-paced setting of an operating room. Efficient frame pruning works by identifying and discarding redundant frames that do not contribute new information, thereby optimizing resource utilization without compromising segmentation accuracy \citep{gao2022rgb}. This approach allows the model to focus computational efforts on critical moments within the surgical video, improving both speed and efficiency. Researchers are also exploring the integration of motion detection algorithms and temporal consistency models to further enhance the performance of SAM in real-time applications \citep{twinanda2016endonet}. These advancements make it feasible to deploy sophisticated segmentation models in surgical settings, facilitating better surgical navigation and instrument tracking and potentially improving patient outcomes.

\begin{table*}[hbtp]
\centering
\caption{Comprehensive summary of tool navigation datasets (Chronologically Sorted)}
\label{tab:datasetcomp}
\resizebox{\textwidth}{!}{%
\begin{tabular}{@{}p{0.8cm}p{2.5cm}p{2.5cm}p{2.5cm}p{1.5cm}p{1.8cm}p{3cm}p{3.2cm}@{}}
\toprule
\textbf{Year} & \textbf{Collection} & \textbf{Data Volume} & \textbf{Surgical Type} & \textbf{Access} & \textbf{Instrument Type} & \textbf{Label Types} & \textbf{Operational Tasks} \\ \midrule
2015 & FetalFlexTool & 21 Videos & Fetal Surgery & Public & Rigid & Bounding-box & Detection \\
2015 & NeuroSurgicalTools & 2476 Images & Neurosurgery & Public & Robotic & Bounding-box & Detection \\
2015 & EndoVis 15 & 9K Images & Colorectal Surgery & Public & Rigid & Pixel-wise, 2D pose & Segmentation, Tracking \\
2016 & Cholec80 & 80 Videos & Cholecystectomy & Public & Rigid & Bounding Box & Detection, Activity, Skills \\ 
2016 & M2CAI16-tool & 16 Videos & Cholecystectomy & Public & Rigid & Tool presence & Detection, Phase \\ 
2017 & Hamlyn & 2 Phantom & Cardiac & Public & NA & Depth map & Tracking \\ 
2017 & ATLAS Dione & 8 Videos & In-vitro Experiments & Public & Robotic & Bounding Box & Detection, Activity \\ 
2017 & EndoVis 17b & 10 Videos & Porcine & Public & Robotic & Pixel-wise & Segmentation, Binary, Parts \\
2018 & m2cai16-tool locations & 16 Videos & Cholecystectomy & Public & Rigid & Bounding-box & Detection \\ 
2018 & LapGyn4 & 55K Images & Gynecologic Surgery & Public & Rigid & No annotation & Multiple \\ 
2018 & EndoVis 18d & 14 Videos & Nephrectomy & Public & Robotic & Pixel-wise mask & Scene Segmentation \\ 
2019 & SCAREDf & 27 Videos & Porcine & NA & NA & Depth + mask & Depth estimation \\ 
2019 & Cata7 & 7 Videos & Cataract Surgery & Private & Rigid & Pixel-wise mask & Segmentation, 3D Reconstruction \\ 
2019 & UCL & 16016 Synthetic Images & Sigmoid resection, colonoscopy & NA & NA & Depth map & Depth estimation \\
2019 & ROBUST-MIS19 & 30 Videos & Proctocolectomy, rectal & Public & Rigid & Instances & Segmentation(Binary,Parts) \\ 
2020 & LapSig300 & 300 Videos & Colorectal Surgery & Private & Rigid & Pixel-wise mask & Segmentation, Parts, Action Recognition \\ 
2020 & Sinus Surgery-L & 3 Videos & Sinus-Live & Public & Rigid & Pixel-wise mask & Segmentation, Phase, Action Recognition \\ 
2020 & Sinus Surgery-C & 10 Videos & Sinus-Cadaver & Public & Rigid & Pixel-wise mask & Segmentation(Binary) \\ 
2020 & UCL dVRK & 20 Videos + Kinematic Data & Ex-Vivo & Public & Robotic & Camera parameters & Segmentation(Binary) \\ 
2020 & HeiSurF & 24 Videos, 9 Test Surgeries & Laparoscopic gallbladder resections & Public & Rigid & Segmentation masks, Instrument classes & Full scene segmentation, Workflow analysis \\
2021 & KvasirInstrument & 590 endoscopic frames & Gastrointestinal Endoscopy & Public & Rigid & Binary masks, Bounding boxes & Tool Segmentation \\ 
2021 & RoboTool & 514 Video Frames, 14720 Synthetic Images & Various Surgical & Public & Robotic & Binary labels, segmentation masks & Synthetic dataset creation \\ 
2021 & I2I Translation & 200,000 Images & Laparoscopic Surgery & NA & Rigid & Style transformation masks & Image style translation \\ 
2021 & SCARED & 9 Datasets & Porcine & Public & Robotic & Depth & Depth estimation \\ 
2021 & CaDTD & 50 Videos & Cataract Surgery & Private & Rigid & Bounding-box & Detection \\
2021 & dVPN & 48702 Images & Nephrectomy & NA & NA & TP, A, SC, Ph & Detection, Action Recognition \\ 
2021 & EndoVis 21 & 33 Videos & Cholecystectomy & Public & Rigid & TP, A, SC, Ph & Detection, Phase, Action Recognition \\ 
2021 & AutoLaparo & 21 Videos & Laparoscopic Hysterectomies & Public & Rigid & Annotated for uterus and instruments & Workflow recognition, Motion prediction, Anatomy segmentation \\ 
2021 & ART-Net & 29 Procedures & Laparoscopic Hysterectomies & Public & Non-robotic & Binary segmentation, tool presence & Instrument segmentation, geometry annotation \\
\bottomrule
\end{tabular}
}
\end{table*}

\subsection{Endoscopic and Laparoscopic Tool Detection Datasets}
The MICCAI Endoscopic Vision (EndoVis) challenge initiated in 2015 have played a crucial role in propelling AI research in endoscopic and laparoscopic tool detection. These challenges offer datasets for diverse tasks like segmentation, tracking, and classification. The availability of high quality datasets is vital in driving advances in the field.  Starting with rigid surgical instrument segmentation and tracking in 2015 \citep{bouget2015detecting}, subsequent challenges expanded to include binary and multi-class instrument segmentation, and by 2018, comprehensive scene segmentation tasks involving both robotic and non-robotic tools as well as anatomical structures were introduced \citep{allan20202018}. The 2019 Robust-MIS challenge elevated this with a dataset of 30 surgical procedures aimed at improving robustness and generalization in binary and instance segmentation of surgical tools \citep{ross2021comparative}. Additional focuses in recent years have included depth estimation in the SCARED sub-challenge, as well as domain adaptation and workflow recognition in the 2020 and 2021 challenges, respectively.
The M2CAI16 Challenge presents two datasets for surgical workflow and tool detection: the m2cai-tool dataset \citep{twinanda2016endonet} annotating the presence of seven surgical tools that feature 15 cholecystectomy videos from the University Hospital of Strasbourg where 10 are used for training and 5 for testing purpose. This dataset was later expanded to m2cai-tool-locations \citep{jin2018tool}, adding spatial annotations for 2,532 frames to aid tool localization. The Cholec80 dataset, developed by Twinanda et al. \citep{twinanda2016endonet}, includes 80 cholecystectomy videos by 13 surgeons, with 86,000 annotated images for tool presence and surgical phase recognition, including tool bounding boxes for 10 videos. It was extended by the ITEC Smoke Cholec80 Image dataset \citep{aksamentov2017deep}, which adds 100,000 frames annotated to differentiate smoke from non-smoke scenarios, aiding research on smoke removal in surgical environments. The Kvasir-Instrument dataset \citep{jha2021kvasir} comprises 590 endoscopic frames from gastroscopies and colonoscopies, providing binary segmentation masks and bounding boxes for various instruments at resolutions from 720×576 to 1280×1024. Additionally, the ART-Net dataset \citep{hasan2021detection} includes 29 laparoscopic hysterectomy procedure videos recorded in 1920×1080 resolution, featuring binary segmentation and geometric data for non-robotic instruments, enhancing its application in tool presence detection.
All these details of the datasets are summarized in Table \ref{tab:datasetcomp}.

\section{ML Applications in Surgical Workflow Analysis}
\label{sec:workflow}
Surgical Workflow Analysis critically employs ML to enhance the understanding and efficiency of surgical procedures by automating the recognition of phases and steps within surgeries. This analysis is key to improving efficiency, safety, and quality in surgical settings. Recent advancements in DL have significantly strengthened the ability to perform nuanced analyses of complex surgical activities. A detailed review by Demir et al. \citep{demir2023deep} highlights the crucial role of DL techniques in recognizing surgical phases and steps, employing advanced architectures such as CNNs, RNNs, and transformers to process sequential video data for identifying critical surgical actions and transitions. These models are particularly effective in complex environments rich in visual data.

The efficacy of these models in surgical workflow analysis is rigorously evaluated using benchmarks such as the HeiChole benchmark, which provides a standardized dataset from cholecystectomy procedures to validate and compare algorithms \citep{wagner2023comparative}. Such comparative analyses are vital for identifying superior computational approaches and fostering ongoing improvements in the field. In addition, there is an increasing emphasis on the application of these advanced analytical methods in specific surgical specialties, including open orthopedic surgery. This involves analyzing intraoperative video data to understand workflows, crucial for training, planning, and providing real-time operational support. Adapting DL models to the unique visual and procedural nuances of specialties like orthopedic surgery presents significant challenges, given the specialized instruments and complex movements involved.

\begin{takeaways}
The field has shifted from CNN/YOLO to transformer/FMs, with promptable segmentation methods that noticeably stabilize predictions across frames. Data curation still dominates results where datasets like EndoVis/ROBUST-MIS and newer sets drive progress. Semi/weak supervision (e.g., SegMatch) reduces annotation load, while FM adapters balance accuracy and efficiency. 
\end{takeaways}

\subsection{Phase Recognition}
Surgical phase recognition is a key aspect of surgical process modeling, with the aim of automatically identifying different stages of a procedure from video data. This capability is crucial to improving surgical training, automating documentation, and supporting real-time decision making in operating rooms. Recent advances in deep learning have notably advanced the accuracy and real-time functionality of phase recognition systems. Machine learning models, especially those employing temporal classification networks like LSTM networks, effectively recognize and segment different surgical phases from video data, structuring the surgical workflow to facilitate relevant information and tool availability at each stage and improving surgical efficiency and safety assessments.  \citep{twinanda2016endonet} introduced EndoNet, a DL architecture that performs phase recognition in surgical scenarios, demonstrating its utility in automating and improving understanding of surgical procedures.


A systematic review by \citep{garrow2021machine} examined the transition from traditional ML methods to advanced DL models in surgical phase recognition, tracing the evolution of increasingly sophisticated algorithms that utilize extensive datasets and complex model architectures \citep{padoy2012statistical}. This progression has significantly enhanced accuracy and real-time applicability. Early methods involved simpler classifiers like Support Vector Machines (SVM) and Hidden Markov Models (HMM), but these were often constrained by their linear decision boundaries and basic assumptions about data sequence dependencies \citep{padoy2012statistical}. The early advancements techniques favor DL methods, particularly CNNs, and RNNs, which effectively manage the spatial and temporal complexities of surgical videos \citep{twinanda2016endonet, zia2021surgical}. Moreover, the development of multi-stage temporal convolutional networks (TCNs), such as the ``Tecno'' model, offers a robust solution that bypasses the limitations of recurrent architectures by using temporal convolutions to capture sequential dependencies more efficiently \citep{czempiel2020tecno, lea2017temporal}. These TCNs are particularly adept at processing sequential data, making them ideal for real-time applications that require instant feedback and recognition during surgical procedures \citep{czempiel2020tecno}. The comprehensive and fine understanding of these models significantly improves the automation of surgical documentation and aids in training surgical residents by offering objective and consistent phase identification \citep{lea2017temporal}.

Where prior work often traded off temporal reach for tractability, some recent works have focused on data efficiency and long-range memory handling. SemiVT-Surge \citep{li2025semivt} introduces a semi-supervised video transformer that integrates temporal consistency regularization with prototype-guided contrast, leveraging large pools of unlabeled OR video to approach fully supervised accuracy on RAMIE \citep{li2024benchmarking} and Cholec80 \citep{twinanda2016endonet} while using a fraction of labels. In parallel, MoSFormer \citep{ding2025mosformer} augments sliding-window transformers with a `Memory of Surgery' that encodes the context of long-term procedure alongside short-term impressions, reducing fragmented predictions on long, branching workflows. Together, such models are demanding as they learn from what hospitals actually have (mostly unlabeled video) while reasoning over minutes rather than seconds.

Besides individual interactions, structured interaction modeling has also matured in the literature. TriQuery \citep{yao2025triquery} casts triplet recognition as query-based decoding with a temporal Top-K update that stabilizes predictions across frames, reports gains on CholecT45, and offers interpretable attention over tools/targets. Similarly, \citep{pei2025instrument} pursue instrument and tissue-guided triplet detection with textual–temporal features, which shows the coupling between localization and interaction semantics. In addition, \citep{chen2025prostatd} proposed ProstaTD  which is a data set of multi-institution robot-assisted prostatectomy with 60,529 frames and 165,567 annotated triplet instances and precise temporal boundaries, addressing the known limitations of single-site coarsely labeled resources such as CholecT50.

Furthermore, \citep{yuan2024hecvl} introduce a novel framework named HecVL, aimed at enhancing surgical computer vision systems through hierarchical video-language pretraining. This method utilizes multi-level video-text pairs from surgical videos---clip-level for specific actions, phase-level for broader activities, and video-level for overall procedural summaries. These pairs enable the HecVL framework to learn rich multi-modal representations without manual annotations. It also employs a fine-to-coarse contrastive learning strategy, creating separate embedding spaces for each hierarchical level, which effectively separates and aligns visual and textual data. The model has been tested across various datasets, where it demonstrates robust capabilities in zero-shot surgical phase recognition, showing strong performance in identifying surgical phases without direct training on those specific phases. This zero-shot learning ability, along with its adaptability to different medical centers marks HecVL as a significant advancement in surgical AI and offers a scalable solution that enhances the generalization and application of surgical computer vision technologies.

\subsection{Tissue and Organ Segmentation}
Tissue and Organ Segmentation is vital in medical imaging, enabling precise identification and delineation of anatomical structures for presurgical planning and intraoperative guidance. DL models, especially U-Nets and Fully Convolutional Networks (FCNs), have dramatically improved the segmentation of medical images, offering high accuracy and real-time processing \citep{ronneberger2015u, long2015fully}. The V-Net architecture, a three-dimensional variant of the U-Net developed by \citep{milletari2016v}, excels in volumetric segmentation, processing entire data volumes simultaneously to effectively capture spatial hierarchies of tissues. The use of volumetric convolutions and skip connections in this model helps to maintain segment boundary accuracy during up-sampling, which is crucial for surgeries requiring high precision. Furthermore, recent enhancements incorporate Generative Adversarial Networks (GANs) and attention mechanisms, which improve the model's ability to differentiate between similar tissues and enhance boundary contrast in complex scenarios \citep{oktay2018attention}.

\subsection{Anatomical Structure Recognition}
Anatomical Structure Recognition is an essential component of advanced surgical systems, focusing on the accurate identification and classification of anatomical structures within the surgical field. This capability is crucial to improving the functionality of automated surgical systems and providing surgeons with real-time decision support. Using advanced ML techniques, particularly CNNs, these systems can efficiently recognize and label complex anatomical features, helping to improve the precision and safety of surgical interventions. \citep{oh2024real} demonstrated the effectiveness of DL models in real-time recognition of anatomical structures, significantly improving both the accuracy and safety of surgical procedures. These models are trained on large datasets of annotated images, allowing them to develop a nuanced understanding of various anatomical nuances that are critical during operations.

 \subsection{Action and Task Recognition}

Action and Task Recognition in surgery identifies specific actions and tasks within phases, crucial for training, performance evaluation, and protocol adherence. Advanced ML techniques, such as CNN and sequence modeling, analyze surgical videos to detect subtle surgeon activities, thus enhancing the monitoring and standardization of practices \citep{brandao2018towards}. Some of the recent developments include a computer vision platform that uses ML algorithms to recognize actions and highlight crucial events during surgeries such as laparoscopic cholecystectomy, contributing to procedural standardization and training improvement \citep{mascagni2021computer}. Additionally, gesture recognition technologies in robotic surgery improve interactions between surgeons and robotic systems, enhancing efficiency and outcomes by providing intuitive control of surgical robots \citep{van2021gesture}.

Autonomous Instruments Control uses ML to automate the control of surgical instruments, improve precision, and minimize human errors by replicating expert maneuvers, offering consistency in intricate tasks, and reducing surgeon fatigue \citep{kassahun2016surgical, shademan2016supervised}. Anomaly Detection and Safety Monitoring employs ML to identify deviations from standard procedures, enhancing safety with real-time alerts and interventions, which are essential in high-stakes environments \citep{wagner2023comparative, maier2018surgical}. It is imperative that these alerts do not overload clinical staff, resulting in  ``notification fatigue". Striking a balance is key. 

Augmented Reality (AR) and Navigation Assistance technologies use ML to overlay critical information directly on the surgical field and provide navigational signals during procedures, significantly impacting surgical planning and execution \citep{bernhardt2017status, cutolo2024augmented, maier2017surgical}. Error detection and feedback mechanisms use ML to monitor surgeries and provide corrective feedback in real time, thereby improving quality, safety and supporting surgical training by providing real-time guidance and post-procedure analysis to beginner surgeons \citep{shademan2016supervised, twinanda2016endonet, lajczak2024md}.

\subsection{Downstream Tasks using Multimodality}
BERT is a foundational language model renowned for its proficiency in understanding and generating human language. In the surgical domain, BERT has been leveraged to improve various aspects of clinical practice, documentation, and decision-making processes. One of the primary applications of BERT in surgery involves tasks of natural language processing (NLP), such as the automatic extraction and interpretation of information from surgical reports, electronic health records (EHR) and operative notes \citep{zhong2024improving}. By accurately parsing complex medical terminology and contextual nuances, BERT-based models facilitate the creation of structured data sets from unstructured textual data, which can be instrumental for clinical research and patient management.

In addition, BERT has been integrated into decision support systems to help surgeons make informed decisions during procedures. For example, by analyzing historical surgical data and outcomes, BERT models can provide predictive insights and recommendations tailored to specific patient cases, thus improving surgical planning and risk assessment \citep{li2023interpretable}. Furthermore, BERT improves communication within surgical teams by enabling more efficient information retrieval and summarization, ensuring that all team members have access to relevant and up-to-date patient information in real-time. This  approach to key information enables high performance within the team setting. 

In the realm of surgical training and education, BERT-powered applications offer personalized learning experiences for surgical trainees. These applications can analyze trainee performance reports, identify areas for improvement, and suggest targeted educational resources, thus fostering continuous professional development \citep{ray2024large}. Furthermore, BERT facilitates the development of intelligent virtual assistants that can support surgeons by answering questions, providing procedural guidelines, and managing administrative tasks, ultimately contributing to increased operational efficiency and reduced cognitive load during surgeries.

LLaMA is renowned for its natural language processing prowess and is enhancing surgical and healthcare operations through a series of specialized systems. LLaVA-Surg, as detailed by \citep{li2024llava}, uses LLaMA to integrate and interpret multimodal data from surgical procedures. This system helps surgical teams by providing dynamic support during operations, helping to identify crucial procedural steps and potential deviations, thus improving surgical safety and efficiency. Following this, Surgical-LLaVA focuses on deepening the understanding of surgical scenarios. Jin et al. \citep{jin2024surgical} highlight how this system utilizes large language and vision models to process complex surgical data, thus improving the precision and responsiveness of surgical interventions.

Furthermore, LlamaCare by \citep{sun2024llamacare} leverages LLaMA's capabilities to improve healthcare knowledge sharing. This system improves the dissemination of medical information and best practices throughout the healthcare community, facilitating better communication between healthcare providers and improving patient care through access to updated and comprehensive medical knowledge.

ChatGPT has also shown significant promise as an intraoperative and educational tool in various surgical disciplines, as highlighted in recent studies. Atkinson et al. \citep{atkinson2024artificial} and Araji et al. \citep{araji2024evaluating} demonstrate how ChatGPT helps decision making during complex surgeries such as Deep Inferior Epigastric Perforator (DIEP) flap procedures and enhances learning experiences for surgical clerkships. These studies illustrate ChatGPT's ability to deliver real-time, accurate responses and educational support, improving surgical safety and training outcomes. Extending its applications, the review by Goglia et al. \citep{goglia2024artificial} further emphasizes ChatGPT's role in abdominal and pelvic surgery, showing its effectiveness in preoperative planning, intraoperative decisions, and patient communication. Together, these studies reflect a transformative shift towards integrating AI in surgical practices, underscoring ChatGPT's potential to optimize surgical workflows, precision, and educational protocols while also highlighting the need for ongoing refinement and cautious integration into clinical settings.

\subsection{Workflow Analysis Datasets}
Workflow analysis in surgical settings benefits from a variety of specialized datasets, each contributing unique insights and challenges to the field. The ATLASDione dataset, compiled by Sarikaya et al. \citep{sarikaya2017detection}, includes 99 videos from robotic tasks performed by clinicians, annotated with tool bounding boxes, action types, durations, and surgeon's skill levels. The dVPN data set from Ye et al. \citep{ye2017self} provides 34,320 pairs of stereo images for training and 14,382 pairs for testing, improving stereo vision and depth estimation in robotic surgeries. The AutoLaparo dataset \citep{wang2022autolaparo} contains 21 high-definition videos of laparoscopic hysterectomy procedures, annotated for tool and anatomy segmentation and workflow recognition. The NeuroSurgicalTools data set \citep{bouget2015detecting} offers 14 neurosurgery videos with detailed instrument annotations, addressing visual challenges like occlusions and reflections. The FetalFlexTool dataset by García-Peraza-Herrera et al. \citep{garcia2017real} provides ex-vivo fetal surgery images and videos, annotated for tool segmentation under varied conditions.

Additional synthetic and simulated data sets include the UCL data sets \citep{rau2019implicit, colleoni2020synthetic}, offering synthetic images and videos rendered to simulate various surgical scenarios. The Laparoscopic Image-to-Image Translation dataset \citep{pfeiffer2019generating} includes 100,000 images derived from CT scans, adapted to various visual styles for model testing. The Sinus Surgery Dataset \citep{qin2020towards} features videos of sinus surgeries annotated for tool segmentation in complex environments. Cata7 \citep{ni2019raunet} focuses on cataract surgery, providing high-resolution images with detailed annotations, while the CaDTD data set \citep{jiang2021semi} extends the CATARACTS data set \citep{al2019cataracts} with semi-supervised learning techniques for cataract surgeries. Lastly, the RoboTool dataset \citep{garcia2021image} includes real and synthetic images for training in robust tool segmentation. These diverse data sets collectively improve the development and testing of machine learning models in various surgical specialties and conditions.

\begin{takeaways}
Temporal and vision-language models are now state-of-the-art for phase and triplet recognition, surpassing recurrent baselines. Graph fusion with kinematics improves robustness, and prompt-tuned CLIP-style encoders offer strong representations with little supervision. Evaluation must reflect long, branching procedures and real OR variability. 
\end{takeaways}

\section{ML Applications in Surgical Training and Simulation}
\label{sec:surgical_training}

ML and DL are significantly enhancing surgical training and simulation, offering profound insights and enhancements in the understanding of surgical scenes. These technologies help create realistic simulations of surgical procedures, allowing trainees to practice and learn in a safe environment. ML and DL tools generate detailed visuals and provide information about patient anatomy and how surgical tools interact with tissues. This helps surgeons understand and navigate the surgical environment better, making it easier for them to perform successful operations. Overall, using ML and DL in surgical training helps improve a surgeon's skills and knowledge, which is crucial for effective and safe surgery. Following are some of the key components that correspond to designing efficient tools and techniques.

\subsection{Key Frame Extraction}

The extraction of key frames from surgical videos is a critical process for efficient video analysis, allowing the identification of important moments without the need to review entire video sequences. An advanced method, proposed by Ma et al. \citep{ma2020keyframe}, uses a diverse and weighted dictionary selection algorithm to identify key frames based on their representativeness and uniqueness. This approach ensures that the selected frames capture critical phases of the surgery, enhancing their educational and analytical value. Another technique, described by Tan et al. \citep{tan2024large}, leverages large-scale DL models to extract sequential key frames, accurately summarizing surgical procedures. This method emphasizes the extraction of key stages of the surgery, providing a concise and informative visual summary, particularly useful for surgical training and procedural reviews.

\subsection{Tissue Classification}
Tissue Classification is crucial in surgical procedures for tasks such as cancer excision, involving the differentiation of tissue types based on characteristics such as appearance, texture and morphology. Machine learning, particularly DL-based techniques, excels in analyzing surgical images and videos to classify tissues with high accuracy, aiding surgeons in making decisions during complex interventions. Cekic et al. \citep{cekic2024deep} introduced a DL approach using CNNs to classify tissue types in surgical videos in real time, effectively distinguishing between healthy and pathological tissues. This precision is vital in oncological surgeries to ensure complete removal of malignant cells while conserving healthy tissue. The field has seen further advancements by integrating more sophisticated architectures like Residual Networks (ResNets) and Inception networks. These networks provide deeper and more complex models that capture a broader range of features from surgical images, significantly enhancing the accuracy of classification \citep{hu2024advancing}.


\subsection{Depth Estimation and 3D Reconstruction}
Depth Estimation and 3D Reconstruction technologies play a crucial role in providing three-dimensional perspectives of the surgical field from two-dimensional images or video feeds, particularly in MIS. These technologies, leveraging ML models such as stereo vision algorithms and structured light approaches, enhance depth perception and spatial understanding, thus improving the precision and safety of surgical interventions. Guni et al. \citep{guni2024artificial} highlighted the effectiveness of ML techniques in producing accurate 3D reconstructions that significantly aid in planning and executing complex surgical procedures. These models are trained on large datasets and identify and interpret depth signals to create detailed 3D maps of the operative area. Recent advancements include using DL frameworks like CNNs and Generative Adversarial Networks (GANs) to improve the accuracy and detail of 3D reconstructions. CNNs process large amounts of image data to detect crucial edges and contours for depth estimation, while GANs enhance detail in partially obscured areas \citep{huang2024real}. Furthermore, the use of time-of-flight (ToF) cameras and laser triangulation with ML algorithms facilitates real-time 3D reconstruction during surgeries, providing immediate feedback for on-the-fly adjustments, crucial for successful surgical outcomes \citep{baptista2024structured}.

\subsection{Surgical Video Generation}
\begin{figure}[hbtp]
    \centering
    \begin{subfigure}[b]{0.48\columnwidth}
        \centering
        \includegraphics[width=\textwidth]{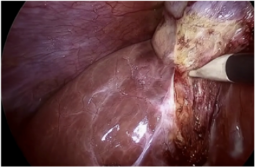}
        \caption{Gallbladder dissection.}
        \label{fig:sub1-Surgen}
    \end{subfigure}
    \hfill
    \begin{subfigure}[b]{0.48\columnwidth}
        \centering
        \includegraphics[width=\textwidth]{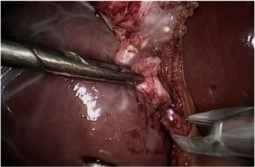}
        \caption{Clipping and cutting.}
        \label{fig:sub2-Surgen}
    \end{subfigure}
    
    \begin{subfigure}[b]{0.48\columnwidth}
        \centering
        \includegraphics[width=\textwidth]{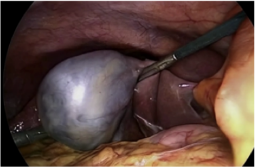}
        \caption{Preparation for removal.}
        \label{fig:sub3-Surgen}
    \end{subfigure}
    \hfill
    \begin{subfigure}[b]{0.48\columnwidth}
        \centering
        \includegraphics[width=\textwidth]{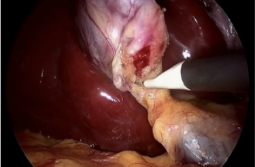}
        \caption{Calot's triangle dissection.}
        \label{fig:sub4-Surgen}
    \end{subfigure}
    \caption{The use of \textit{SurGen} \citep{cho2024surgen}, a \textit{text-guided diffusion model for surgical video synthesis}, to generate key stages of gallbladder surgery. These images highlight the potential of foundation models to improve surgical scene understanding by generating realistic visuals for training and planning. Images adapted from \citep{cho2024surgen}.}
    \label{fig:surgical}
\end{figure}

The generation of realistic surgical video through ML models opens new possibilities for training, simulation, and research. A recent work designed SurGen \citep{cho2024surgen}, an innovative approach to surgical video generation which is a text-guided diffusion model designed to create detailed and accurate surgical videos from textual descriptions. This model leverages the capabilities of diffusion models, which construct images by gradually refining patterns of noise into structured visuals. By integrating text input, SurGen allows users to specify what the video should depict, making it a powerful tool to create customized training materials or to simulate surgical scenarios to study possible outcomes and strategies, as shown in Fig.~\ref{fig:surgical}.

Recent advancements in the field have shown that video generation models can also leverage temporal coherence and multi-modal data to enhance the realism of generated surgical procedures. For example, the work by Wang et al. \citep{wang2024leo} introduced a method for surgical video synthesis that maintains temporal consistency across frames, which is crucial for realistic simulations. Similarly,\citep{yamada2024multimodal} explored the integration of multi-modal inputs such as surgical instrument tracking data, further improving the fidelity of generated videos by aligning visual and kinematic information.

\subsection{Anomaly Detection}
Anomaly detection in surgical settings is critical for ensuring patient safety and enhancing surgical precision. This technology leverages advanced ML techniques to identify deviations from normal procedures, which can indicate potential risks or complications that might occur when the position of the surgical dissection plane alters.

In recent developments, unsupervised learning techniques have been applied to robotic surgery through the use of deep residual autoencoders. These models learn to reconstruct normal surgical activities by training on large datasets of standard operations. During actual procedures, significant deviations from the expected reconstruction are detected and flagged as anomalies, highlighting potential issues. The unsupervised nature of this approach makes it particularly valuable in environments where labeled data is scarce or unavailable, making it an ideal tool for real-time monitoring of complex robotic surgeries \citep{li2024advances}. This method ensures that deviations from expected surgical patterns are identified, improving surgical accuracy and patient safety.

\begin{figure}[h!]
  \centering
  \includegraphics[width=\columnwidth]{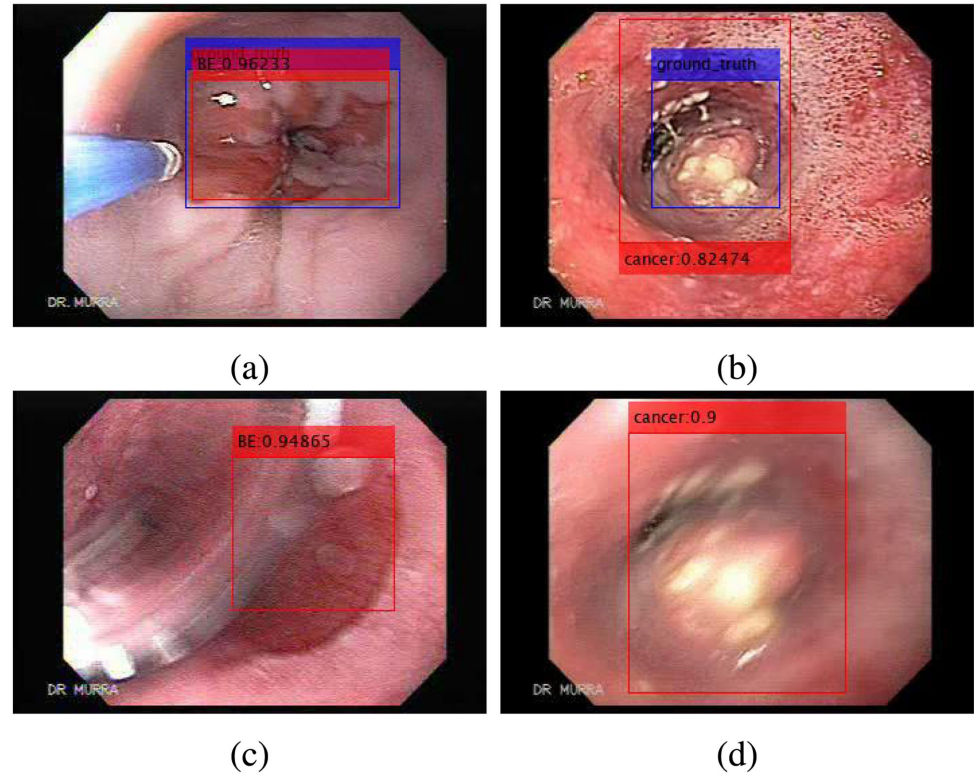} 
  \caption{Detection results from complex endoscopic frames, highlighting challenges in identifying esophageal abnormalities in surgical scenes: (a) tool occlusion, (b) bubbles, (c) motion blur, and (d) fog. Red boxes indicate AI model predictions, and blue boxes represent ground truth annotations. Images from \citep{ghatwary2020learning}.}
  \label{fig:endo} 
\end{figure}

This method focuses on capturing temporal and spatial abnormalities in endoscopic videos of the esophagus, as shown in Fig.~\ref{fig:endo}. By employing CNNs that track changes over time, the system can detect subtle anomalies that may be indicative of esophageal diseases, such as early-stage cancers or dysplasia \citep{ghatwary2020learning}. This method demonstrates how ML can improve diagnostic precision in challenging clinical contexts.



\subsection{Surgical Video Summarization}

Surgical video summarization is crucial in the medical field for education, documentation, and preoperative planning, enabling efficient review of lengthy and complex surgeries. ML and DL techniques play a vital role in this process by identifying key events, actions, and phases in surgical videos to create concise summaries without losing essential information. One prominent method is the automatic summarization of endoscopic surgical videos, where systems like those developed by \citep{king2022automatic} automatically generate summaries by identifying critical surgical phases, condensing lengthy recordings into essential, manageable segments. This allows surgeons and trainees to quickly review procedures, capturing major events such as the start and end of critical phases. \citep{xiao2019deep} introduced another method using hierarchical DL models that capture both temporal and spatial information to accurately identify important segments. This technique uses a combination of convolutional and recurrent neural networks to provide contextually rich summaries.

 \citep{esteva2019guide} suggest combining video summarization with real-time feedback systems to enhance surgery decision-making, potentially integrating with surgical robots or augmented reality platforms for real-time procedural guidance. Advanced techniques further refine surgical video summarization such as \citep{king2023automatic}, they utilized object detection and Hidden Markov Models (HMMs) to segment and summarize skull base surgeries by highlighting key instruments and actions. In addition, live tags from surgical teams mark important moments, enabling collaborative summarization. A deep multi-scale pyramidal features network dynamically summarizes complex surgeries by capturing hierarchical structures in the content, suitable for procedures requiring multiple detail levels \citep{hashimoto2018artificial}. Furthermore, techniques such as deep feature matching and motion analysis specifically tailor summarization for wireless capsule endoscopy by focusing on areas of pathological interest \citep{garcia2023videosum}. These innovations collectively improve the field of surgical video summarization, providing powerful tools for education, documentation, and clinical review.

\subsection{Surgical Skills Assessment}
Surgical skills assessment is a fundamental aspect of maintaining high standards in surgical training and practice, and recent advancements in ML have led to the development of innovative and objective methods for evaluating these skills. One such approach is the Contrastive Regression Transformer model, which is designed to assess surgical skills during robotic surgeries by analyzing video data. This model captures subtle movements and decision-making processes to provide a quantitative assessment that aids in improving surgical performance \citep{anastasiou2023keep}. Additionally, video-based analysis of recognized surgical gestures and skill levels has been explored in the work by \citep{wang2020towards}, where ML techniques are used to correlate specific actions with skill levels, offering a more detailed and interpretable evaluation of surgical proficiency. This method enhances the feedback given to surgeons during training, ensuring a more tailored and effective learning process. Furthermore, the link between a surgeon's technical skills and patient outcomes has been well-established, as highlighted by \citep{stulberg2020association}, emphasizing the critical impact that high surgical proficiency has on improving clinical outcomes. These developments demonstrate the growing role of ML in refining surgical training and ensuring optimal patient care.

\begin{table*}[ht]
\centering
\caption{Overview of Foundation Models in medical imaging across various modalities and tasks. This table summarizes key methodologies, their model specifications, modalities, downstream tasks, and primary findings. The abbreviations here are Seg: Segmentation, Cls: Classification, Det: Detection, IE: Image Enhancement}
\label{tab:flapplications}
\resizebox{\textwidth}{!}{%
\begin{tabular}{@{}p{1cm}p{3cm}>{\centering\arraybackslash}m{1cm}p{2cm}p{3.1cm}p{2.5cm}p{8cm}@{}}
\toprule
\textbf{Year} & \textbf{Method} & \textbf{Code} & \textbf{Model} & \textbf{Modality} & \textbf{Downstream Task} & \textbf{Key Finding}\\
\midrule
2023 & MA-SAM \citep{chen2024ma} & \textcolor{blue}{\checkmark} & ViT (SAM) & CT, MRI, Endoscopy & Seg & Improved robustness across modalities \\ 
2023 & Endo-FM \citep{wang2023foundation} & \textcolor{blue}{\checkmark} & ViT & Endoscopy & Seg, Cls, Det & Surpasses SOTA with significant margins in pre-training and transfer learning\\
2023 & VisionFM \citep{qiu2023visionfmmultimodalmultitaskvision} & \textcolor{blue}{\xmark} & - & Multimodal images &  Cls &  Delivers expert-level diagnostic accuracy in ophthalmology, surpassing traditional models and specialists with its modality-agnostic approach\\
2023 & Polyp SAM++ \citep{article} & \textcolor{blue}{\checkmark} & ViT (SAM) & Endoscopy & Seg & Advances segmentation using text-guided SAM, enhancing localization and accuracy, effectively handling complex colonoscopy images\\
2024 & SP-SAM \citep{yue2024surgicalpartsam} & \textcolor{blue}{\checkmark} & ViT & Endoscopy & Seg & Integrates part-level prompts with image embeddings, enhancing detailed structure segmentation, with minimal parameters\\
2024 & Polyp-SAM\citep{li2024polyp} & \textcolor{blue}{\checkmark} & ViT (SAM) & Endoscopy & Seg & Utilizes fine-tuned SAM for polyp segmentation, achieving state-of-the-art results with superior generalization, Dice scores \\
2024 & Swinsam\citep{feng2025swinsam} & \textcolor{blue}{\checkmark} & ViT (SAM) & Endoscopy & Seg & Integrates Swin Transformer decoder with SAM encoder, improving segmentation detail significantly, enhancing performance metrics  \\
2024 & SurgicalSAM\citep{yue2024surgicalsam} & \textcolor{blue}{\checkmark} & ViT (SAM) & Endoscopy & Seg & Introduces efficient tuning for SAM with a class prompt encoder, achieving SOTA performance with reduced parameter complexity and improved pipeline simplicity \\
2024 & USFM\citep{jiao2024usfm} & \textcolor{blue}{\xmark} & - & US & Seg, Cls, IE & Demonstrates adaptability across organs and tasks with minimal annotations, enhancing robust feature extraction with spatial-frequency dual masking\\
2024 & LVM-Med\citep{mh2024lvm} & \textcolor{blue}{\checkmark} & ResNet, ViT & Multimodal images  & Seg, Cls, Det & Sets new benchmarks in medical imaging by leveraging large-scale self-supervised learning and second-order graph matching to enhance feature learning and adaptability \\
2024 & VoCo\citep{wu2024voco} & \textcolor{blue}{\checkmark} & Swin & CT  & Seg, Cls &   Utilizes volume contrastive learning to encode high-level semantics, significantly enhancing performance on segmentation and classification tasks\\
2024 & RudolfV\citep{dippel2024rudolfv} & \textcolor{blue}{\xmark} & ViT(DINO-v2) & Pathology  & Cls &  Utilizes pathologist-informed design and semi-automated data curation from a diverse dataset, significantly enhancing diagnostic accuracy and robustness across various benchmarks \\
2024 & AFTerSAM\citep{yan2024after} & \textcolor{blue}{\xmark} & ViT (SAM) & CT  & Seg &  Enhances SAM with Axial Fusion Transformer, improving 3D medical image segmentation by integrating intra-slice and inter-slice contextual information with minimal training data\\
2024 & PUNETR\citep{fischer2024prompt} & \textcolor{blue}{\checkmark} & - & CT  & Seg &  Introduces prompt tuning for efficient semantic segmentation, achieving significant parameter efficiency with robust performance on medical imaging datasets \\

\midrule

\end{tabular}
}
\end{table*}

\subsection{Vision Language Foundation Model}
DINO \citep{caron2021emerging} has shown promise in revolutionizing surgical imaging by enabling improved interpretation and segmentation of medical images through its self-supervised learning approach. In surgical settings, DINO's ability to learn detailed visual representations from unlabeled surgical imagery can significantly improve the accuracy and efficiency of surgical planning and intraoperative guidance. For example, DINO has been applied to automated segmentation of tumor boundaries in real-time during oncological surgeries, helping surgeons achieve more precise excision and margin control. This technology also facilitates the development of advanced diagnostic tools that can automatically differentiate between healthy and pathological tissues, enhancing the surgeon's ability to make informed decisions during complex procedures \citep{koksal2024surgivid}. In addition, by improving the quality and usability of intraoperative images, such as in laparoscopic video feeds, DINO contributes to safer and more efficient surgical workflows. This capability is crucial for MIS procedures, where the clarity and detail of visual information directly impacts precision surgery and therefore reduced postoperative complications. 

Recently, ``Surgical-DINO,'' an adaptation of the DINO model, has been specifically designed to enhance depth estimation in endoscopic surgery, demonstrating the adaptability of Foundation Models to specialized surgical tasks \citep{cui2024surgical}. Moreover, the SURGIVID project leverages DINO's self-supervised learning framework for annotation-efficient surgical video object discovery, further highlighting its utility in reducing the labor-intensive process of video annotation in surgical training and analysis \citep{koksal2024surgivid}. Furthermore, DINO's capabilities could be instrumental in analyzing datasets like EgoSurgery-Tool, which focuses on detecting surgical tools and hands from egocentric perspectives in open surgery videos, potentially enhancing the training and performance of AI models in recognizing and interacting with complex surgical environments \citep{fujii2024egosurgery}. The adoption of DINO in these surgical applications not only streamlines the surgical process but also opens possibilities for more adaptive and responsive surgical systems, potentially reducing the incidence of complications and improving overall surgical outcomes\citep{rabbani2024can}.

CLIP and other vision-language Foundation Models (VLMs) are also reshaping the landscape of medical imaging by effectively linking textual descriptions to visual data. This capability is pivotal for diagnosing radiological images, enhancing surgical planning, and facilitating educational initiatives. The zero-shot learning capability of CLIP introduced by \citep{radford2021learning} allows it to recognize and categorize medical conditions across various imaging modalities, such as X-rays and MRIs, without direct training on specific medical datasets. This cross-modal understanding is especially beneficial in scenarios where large annotated datasets are scarce. Building on this application, \citep{kerdegari2024foundational} demonstrated how Foundation Models could be adapted to improve the detection and categorization of pathologies in specialized medical imagery, such as endoscopy images for the diagnosis of gastric inflammation (gastritis). This study illustrates CLIP's potential to enhance diagnostic accuracy in complex imaging scenarios.

Further expanding the scope of Foundation Models in healthcare,  \citep{sun2024medical} provides a comprehensive examination of medical multimodal foundation models in clinical diagnosis and treatment. Their research focuses on various applications, challenges, and future directions of these technologies in healthcare. They highlight how these models integrate diverse data types by combining visual, textual, and possibly even genomic information to offer more nuanced diagnostics and treatment strategies. This integration facilitates a deeper understanding of patient data, crucial for personalized medicine and advanced treatment planning.  \citep{zhao2023clip} also contribute to this body of knowledge by surveying the deployment of CLIP in medical fields, revealing its transformative impact from routine diagnostics to complex surgical interventions. Table \ref{tab:flapplications} provides a detailed overview of these Foundation Models, outlining their specifications, modalities, and key advancements in enhancing precision and robustness across various medical imaging and clinical tasks.

Moreover,  \citep{yuan2023learning} introduce an innovative approach called SurgVLP (Surgical Vision Language Pre-training). This method enhances surgical computer vision by learning from surgical video lectures without needing manual annotations, which are often labor-intensive and not scalable. It addresses a significant gap in existing models that rely solely on visual input and limited data sets, limiting their ability to generalize across varied surgical procedures and tasks. SurgVLP exploits the untapped educational content in surgical e-learning platforms by using automatic speech recognition (ASR) systems \citep{bhardwaj2022automatic} to transcribe video lectures into text. This process creates a rich dataset of video and corresponding textual data, encompassing detailed descriptions of surgical actions, tools, and anatomical references. By integrating two types of ASR systems, i.e., Whisper \citep{radford2023robust} for general language and AWS Medical Transcribe for medical-specific terms, SurgVLP captures a broad range of linguistic nuances which is essential for accurate multi-modal learning.

The core of SurgVLP's methodology is a novel contrastive learning objective that aligns video clips with their corresponding text transcriptions in a shared latent space. This alignment enables the model to effectively synthesize visual and textual information, enhancing its interpretative and analytical capabilities. SurgVLP's practical applications are demonstrated through its ability to adapt to new surgical tasks without specific fine-tuning. The model has been rigorously tested on several tasks, including vision-and-language tasks, where it performs functions like video retrieval and generating surgical reports based on textual queries, and vision-only tasks, such as recognizing surgical tools and phases, where it also shows promising results. These capabilities suggest that SurgVLP can significantly improve AI-assisted systems in operating rooms, making them more intelligent and responsive. 

Moreover, \citep{yuan2025procedure} also introduce the PeskaVLP framework which is designed to further enhance surgical video-language pretraining. This framework addresses key challenges such as textual information loss and spatial-temporal alignment difficulties in multi-modal data sets. PeskaVLP incorporates hierarchical knowledge augmentation by using large language models to refine and enrich textual descriptions derived from surgical lectures. This method ensures that language supervision remains both accurate and comprehensive which reduces the risk of overfitting and enhancing the text's quality. A significant feature of PeskaVLP is its use of a dynamic time warping (DTW) based loss function, which effectively aligns video frames with their corresponding textual descriptions. By embedding procedural knowledge directly into the pretraining process, PeskaVLP maintains essential surgical concepts and meanings and thus improves the model's robustness and transferability. In terms of applications and performance, PeskaVLP has shown superior capabilities in zero-shot transferability across various public datasets for surgical scene understanding and cross-modal retrieval tasks. The framework significantly improves the alignment and understanding of multi-modal data, which is essential for accurately recognizing surgical phases and enhancing real-time decision-making in surgical environments. 
In addition, SurgVISTA \citep{yang2025large} proposes the first video-level pretraining for surgery by jointly modeling spatial structures and temporal dynamics on 3,650 videos (~3.55M frames) across >20 procedures, and then shows consistent gains over both natural-image and surgery-image pretraining across 13 benchmarks (phases, actions, segmentation). Complementing this, SurgVLM \citep{zeng2025surgvlm} builds SurgVLM-DB (~1.81M images and 7.79M conversation pairs spanning 10 tasks) and introduces SurgVLM-Bench, finding that domain-specialized VLMs (ranging from 7B to 72B) substantially outperform general VLMs in surgical perception, temporal analysis, and reasoning. 
\subsection{Visual Question Answering}

\begin{table*}[h!]
\centering
\caption{Summary of recent advancements in Visual Question Answering (VQA) for surgical applications.}
\label{tab:vqa_summary}
\resizebox{\textwidth}{!}{%
\begin{tabular}{p{0.25\linewidth} p{0.36\linewidth} p{0.16\linewidth} p{0.14\linewidth} p{0.25\linewidth}}
\toprule
\textbf{Paper} & \textbf{Key Findings} & \textbf{Dataset Used} & \textbf{Model Type} & \textbf{Application Area} \\
\midrule

SurgVLM \citep{zeng2025surgvlm} & Foundation surgical VLM with \textit{SurgVLM-DB} (large-scale surgical video–text) and \textit{SurgVLM-Bench}; covers perception, temporal reasoning, and VQA. & SurgVLM-DB & Foundation VLM (Qwen2.5-VL based) & Multi-task incl. VQA \\

Challenging VLMs with Surgical Data \citep{mayer2025challenging} & Creates a surgical QA dataset with typed questions; profiles strengths/limits of general VLMs under surgical distribution shift. & Surgical QA (new) & Evaluation dataset \& study & Capability profiling \\

Surgical-VQLA++ \citep{bai2025surgical} & Improves robustness for localized answering via adversarial contrastive learning. & Surg-LA & Contrastive V-L & Error detection / localization \\

SurgXBench \citep{surgxbench} & Explainable VLM benchmark for surgery; emphasizes grounded answers and attribution for surgical QA. & SurgXBench & Benchmark (explainability) & VLM evaluation / surgical VQA \\

LLaVA-Surg \citep{li2024llava} & Builds \textit{Surg-QA} (102k video–instruction pairs) and trains a surgical assistant VLM. \textit{(recent)} & Surg-QA & Vision–Language (LLaVA) & Video QA / assistant \\

GP-VLS \citep{schmidgall2024gp} & General-purpose surgical VQA across comprehensive datasets; emphasizes broad evaluation. & GP-VLS & General V-L Model & Surgical evaluation \\

SSG-VQA-Net \citep{yuan2024advancing} & Scene-graph guided reasoning improves VQA by encoding geometric/relational structure. & SSG-QA & Scene Graph Network & Scene understanding \\

Surgical-LVLM \citep{wang2024surgical} & Adapts Qwen-VL to surgical grounded VQA with VP-LoRA; improves spatial grounding and answering on robotic surgery data. & Surg-LVLM & Adapted VLM (Qwen-VL) & Complex surgical environments; Grounded surgical VQA \\

Surgical-LLaVA \citep{jin2024surgical} & Tailors LLaVA for conversational multi-modal assistance in surgery (instruction tuning on surgical data). & LLaVA-Surg & LLaVA-based & Multimodal assistance \\

Surgical-VQLA \citep{bai2023surgical} & Gated vision–language embeddings for localized answers in robotic surgery. & Robo-Surg & Gated V–L Embedding & Robotic surgery \\

CAT-ViL \citep{bai2023cat} & Co-attention gated model enabling efficient training for robotic surgery VQA. & Surg-CAT & Co-Attention V–L & Surgical training \\

LV-GPT \citep{seenivasan2023surgicalgpt} & Improved word–vision token sequencing for VQA in surgical settings. & Surg-QA & Language–Vision GPT & Robotic surgery assistance \\


Surgical VQA \citep{seenivasan2022surgical} & Introduced the Surgical-VQA task and datasets; vision–text transformer baselines for classification and sentence answers. & Surg-VQA & Transformer-based & Surgical scene QA \\


\bottomrule
\end{tabular}%
}
\end{table*}

Visual Question Answering (VQA) in surgery uses AI to analyze visual data from surgical videos, enhancing decision-making and educational experiences for medical professionals. This application interprets dynamic visual content in real time, deepening understanding of complex surgical procedures, and serves as a crucial tool for training and operational assistance. The Surgment system pioneers this field with advanced segmentation techniques, enabling specific queries about visual elements in surgical videos, such as identifying tools or recognizing procedural actions, thereby enriching video-based surgical training \citep{wang2024surgment}. The Surgical-VQA method utilizes transformer architectures, adept at processing video streams, to provide precise, contextually relevant answers, enhancing real-time surgical decision-making \citep{seenivasan2022surgical}.

Further extending these capabilities, the Language-Vision GPT (LV-GPT) model integrates visual data processing with traditional GPT-2, using a vision tokenizer and vision token embeddings designed for surgical VQA. This model outperforms existing frameworks on datasets like the Endoscopic Vision Challenge Robotic Scene Segmentation 2018 and CholecTriplet2021, setting new benchmarks for AI-driven VQA applications in surgery \citep{seenivasan2023surgicalgpt}. Furthermore, VQA technologies significantly aid postoperative analysis, allowing surgical teams to review complex procedures with AI-enhanced visual data analysis, potentially improving surgical techniques and outcomes. Integrating VQA with augmented reality (AR) and virtual reality (VR) technologies could further revolutionize surgical training and operative workflows by providing AI-generated annotations and insights during procedures, enhancing surgeons' field of view, reducing errors, and improving outcomes. Recent advances in VQA for surgical applications include transformer-based models and sophisticated systems that improve the robustness and precision of VQA tasks in complex scenarios. These developments underscore the increasing importance of VQA systems in surgical training, decision support, and procedural guidance, promising to transform surgical education and operations by improving the cognitive capabilities of surgical systems.

Table \ref{tab:vqa_summary} summarizes recent advancements in VQA for surgical applications, showcasing a variety of models, datasets, and key findings that improve tasks such as scene understanding, error detection, and surgical training. It highlights innovative approaches like transformer-based models and graph networks, which enhance the interpretability and accuracy of surgical VQA systems.

\subsection{Surgical Video Retrieval}
Efficient retrieval of surgical video content is vital in education, training, and clinical review, allowing users to quickly access relevant video segments based on specific queries. One notable method involves unsupervised feature disentanglement, which separates and identifies key features within surgical videos to improve the accuracy of retrieval systems in MIS \citep{wang2022unsupervised}. This unsupervised approach improves the ability to search large volumes of video data (analogous to a comprehensive surgical video library) without relying on extensive labeling, thus streamlining access to critical information during surgical review and training, as also presented in a framework named Surch \citep{kim2023surch}. Surch provides a novel web‐based interface for structural search and comparison of surgical videos that leverages automatically generated procedural graphs to expose the latent step-by-step structure of procedures. It first converts each video into a graph of surgical phases via a CNN-LSTM phase detector and a clustering pipeline that uses custom vectorization and weighting schemes to capture features like recursive steps and unique paths. It then enables users to filter videos by clicking nodes or edges in the graph. This method has been evaluated on 296 robot-assisted radical prostatectomy videos. 
\begin{table*}[hbtp]
\centering
\caption{Comparative analysis of deep learning models for real-time segmentation and instrument identification across surgical procedures. Tasks are grouped and metrics are standardized for readability.}
\label{tab:comparison-models}
\begin{threeparttable}
\renewcommand\arraystretch{1.15}
\setlength{\tabcolsep}{5pt}

\resizebox{\textwidth}{!}{%
\begin{tabular}{@{}l
                >{\raggedright\arraybackslash}p{2.9cm}
                >{\raggedright\arraybackslash}p{2.6cm}
                >{\raggedright\arraybackslash}p{3.0cm}
                >{\raggedright\arraybackslash}p{3.2cm}
                c
                >{\raggedright\arraybackslash}p{2.4cm}
                >{\raggedleft\arraybackslash}p{2.3cm}@{}}
\toprule
\textbf{Ref.} & \textbf{Task} & \textbf{Procedure} & \textbf{Dataset(s)} & \textbf{Model} & \textbf{Modality} & \textbf{Metric} & \textbf{Score} \\
\midrule

\groupheader{Tool segmentation (binary / parts / instance)}
\citep{pakhomov2020searching} & Instrument segmentation (binary, parts) & Porcine procedures & EndoVis 2017 & ResNet18 & \modI & Dice (binary, parts) & 89.6\%, 76.4\% \\
\citep{lee2019weakly} & Instrument segmentation (binary) & Phantom / Porcine & NPA & LinkNet-152 & \modI & Dice & 88.9\% \\
\citep{baby2023forks} & Instrument instance segmentation & MIS (mixed) & EndoVis 2017/2018 & Custom + classifier head & \modI & AP (instance) & +12 AP vs. SOTA (EV2017) \\
\citep{marullo2023multi} & Tool seg. + blood event detection & Laparoscopic & In-house videos & Multi-task CNN (U-Net enc.) & \modV & Dice (tool, blood) & 90.63\%, 81.89\% \\
\citep{yue2024surgicalsam} & Instrument segmentation (prompted) & MIS (mixed) & EndoVis 2017/2018 & SurgicalSAM + class prompts & \modI & IoU / Dice & SOTA (few params) \\

\addlinespace[2pt]
\groupheader{Anatomy segmentation / scene semantics}
\citep{kolbinger2023anatomy} & Anatomy segmentation & Laparoscopic & DSAD (13{,}195 imgs) & DeepLabv3, SegFormer & \modI & IoU (range) & 0.23–0.85 (organ-dependent) \\
\citep{kolbinger2024strategies} & Anatomy segmentation (robustness) & Laparoscopic & DSAD & SegFormer, DeepLabv3 & \modI & IoU, Acc, Prec, Rec, F1, Spec, HD, ASSD & All improved vs. baselines \\
\citep{mao2024pitsurgrt} & Landmarks + anatomy & Pituitary & In-house & PitSurgRT (HRNet) & \modV & IoU ($\Delta$), landmark error & +4.39\% IoU; error $\downarrow$ \\
\citep{park2024deep} & Multi-class semantics & RALP (prostate) & Intraop videos & Reinforcement U-Net & \modV & Dice (per class), IoU & 0.96, 0.74, 0.85, 0.84; IoU 0.77 \\

\addlinespace[2pt]
\groupheader{Tool detection / classification}
\citep{colleoni2019deep} & Instrument detection & Colorectal & EndoVis 2015 & 3D FCNN & \modI & Dice & 85.1\% \\
\citep{liu2020anchor} & Instrument detection & In-vitro & Atlas Dione; EndoVis 2015 & Stacked Hourglass & \modI & mAP (two sets) & 98.5\%, 100\% \\
\citep{wang2019graph} & Tool classification & Cholecystectomy & m2cai16-tool; Cholec80 & 3D DenseNet + GCN & \modI & mAP (two sets) & 90.2\%, 90.13\% \\

\addlinespace[2pt]
\groupheader{Temporal consistency / video stability}
\citep{grammatikopoulou2024spatio} & Spatio-temporal seg. & Lap.; Partial nephrectomy & CholecSeg8k; Private & Spatio-temporal network & \modV & mIoU (gain) & +1.30 pp, +4.27 pp vs. per-frame \\

\addlinespace[2pt]
\groupheader{Referring / intention-aware segmentation}
\citep{wang2024video} & Referring video instrument segmentation & Robotic & In-house videos & VIS-Net + relation graph & \modV & Ref-VIS (IoU/mAP) & $\uparrow$ vs. prior SOTA \\
\citep{chen2024asi} & Audio-driven instrument segmentation & General surgery & EndoVis 2017/2018 & ASI-Seg + SAM & \modI & IoU (global, intention) & 82.37\%, 71.64\% \\

\addlinespace[2pt]
\groupheader{Phases / gestures / tasks (context for segmentation)}
\citep{namazi2019attention} & Phase boundary detection & Cholecystectomy & Cholec80 & LSTM & \modI & MAE (boundary) & 48 s \\
\citep{jin2020multi} & Phase recognition & Cholecystectomy; Colorectal & Cholec80 & VGG-50 + LSTM & \modI & Acc & 89.2\% \\
\citep{petscharnig2018early} & Phase recognition & Gynecology & NPA & GoogLeNet & \modI & AP & 79.6\% \\
\citep{kannan2019future} & Surgery type recognition & Nine surgeries & NPA & VGG16 + LSTM & \modI & Acc & 75\% \\
\citep{aksamentov2017deep} & Surgery time prediction & Cholecystectomy & Cholec120 & ResNet-152 & \modI & MAE (time) & 460 s \\
\citep{nwoye2020recognition} & Fine-grained activities & Cholecystectomy & Cholec80 & ResNet18 & \modI & Acc & 24.8\% \\
\citep{qin2020temporal} & Gesture segmentation & In-vitro & JIGSAWS; NPA & VGG16 & \modI+\modK+\modE & Acc (two sets) & 86.3\%, 89.4\% \\
\citep{zhao2018fast} & Trajectory segmentation & In-vitro & JIGSAWS & Dense CNN & \modI+\modK & mAP & 70.6\% \\

\bottomrule
\end{tabular}%
} 

\vspace{3pt}
\begin{tablenotes}[flushleft]
\footnotesize
\item \textbf{Modality codes:} \modI = images; \modV = video; \modK = kinematics; \modE = events.
\item \textbf{Metrics:} Acc = accuracy; AP/mAP = average precision; Dice/IoU as reported; MAE = mean absolute error; “pp” = percentage points.
\item Where papers report multiple datasets or settings, we list both values (comma-separated) or a qualitative improvement (e.g., “$\uparrow$ vs. SOTA”).
\end{tablenotes}
\end{threeparttable}
\end{table*}

\subsection{Concluding Insights and Implications}

This review has explored a wide range of methodologies and applications in the field of surgical video analysis and instrument detection, as outlined in Table \ref{tab:comparison-models}. The exploration highlights the rapid progress in deploying FMs, such as advanced neural networks, to improve surgical outcomes and support real-time decision-making.

These technologies have evolved from basic instrument classification to sophisticated dynamic phase recognition, showcasing the substantial potential to increase the precision and efficiency of surgical procedures. Adopting state-of-the-art neural architectures and segmentation techniques has particularly enhanced the sensitivity and specificity of these models, making them essential tools in the operating room. This progression demonstrates how Foundation Models not only refining current surgical practices but also setting the stage for future innovations in medical technology.

\subsection{Computational Complexity of FMs in Surgical Applications}

FMs have revolutionized the field of surgical AI by offering robust generalization capabilities and exceptional performance across a variety of tasks \citep{khan2025comprehensive}. However, these models typically require vast amounts of data and considerable computational resources for training, making it challenging for deployment in clinical environments. This section explores effective strategies to mitigate the computational demands of Foundation Models, ensuring their practical applicability in surgical settings.

Foundation Models like SAM \citep{kirillov2023segment} and CLIP \citep{radford2021learning} encompass diverse tasks and modalities, which require extensive training on large datasets. This extensive training enables them to perform well across different tasks, but also makes them resource-intensive. In surgeries, where responses need to be immediate and accurate, the heavy computational demand of these models can be a significant limitation. For this purpose, several innovative techniques have been developed to adapt Foundation Models to be more resource-efficient, making them suitable for real-time surgical applications:

\begin{itemize}
    \item \textit{Adapters:} Adapters are small modules added to a pre-trained model that can be adjusted to new tasks. They require changing only a few parts of the model, which reduces the computational load significantly. In surgery, adapters can help tailor models to recognize specific instruments or actions using limited data without needing extensive retraining.

    \item \textit{Low-Rank Adaptation (LoRA):} LoRA modifies a model's deep structures in a way that reduces the amount of data they handle at once, thereby cutting down on the computing power required. This technique is useful for refining models to perform specific tasks like VQA and grounding for intricate surgical contexts \citep{wang2024surgical}.

    \item \textit{Prompt Tuning: }This method tweaks the inputs given to the model to guide it toward a particular function using the model's existing knowledge base. This approach is computationally light and can be used to adjust models for specific tasks, such as analyzing surgical videos, without extensive reprogramming \citep{wang2023sam}.

    \item \textit{Knowledge Distillation: }This process involves training a smaller, more manageable model to mimic a larger one. The smaller model retains much of the larger model's effectiveness but uses less computational power, making it better suited for use in surgeries where fast processing is crucial \citep{ding2022free}.

    \item \textit{Quantization and Pruning: }These techniques reduce the model's size and speed up its operations. Quantization decreases the precision of the numbers the model uses, and pruning removes parts of the model that have little impact on its performance \citep{kuzmin2023pruning}. Both adjustments help the model run faster and more efficiently, which is essential in a surgical setting.

\end{itemize}

\begin{takeaways}
Synthetic data (diffusion/augmentation), simulation, and VL agents enable scalable training and objective skill assessment. FM-based assistants can summarize, retrieve, and guide, but require uncertainty awareness and guardrails. Linking simulators, outcome data, and OR telemetry will close the sim-to-real gap. 
\end{takeaways}

\section{Open Issues and Future Research Directions in Surgical Scene Understanding}
\label{sec:open-issues}

\begin{figure}[!h]
  \centering
  \includegraphics[width=\columnwidth]{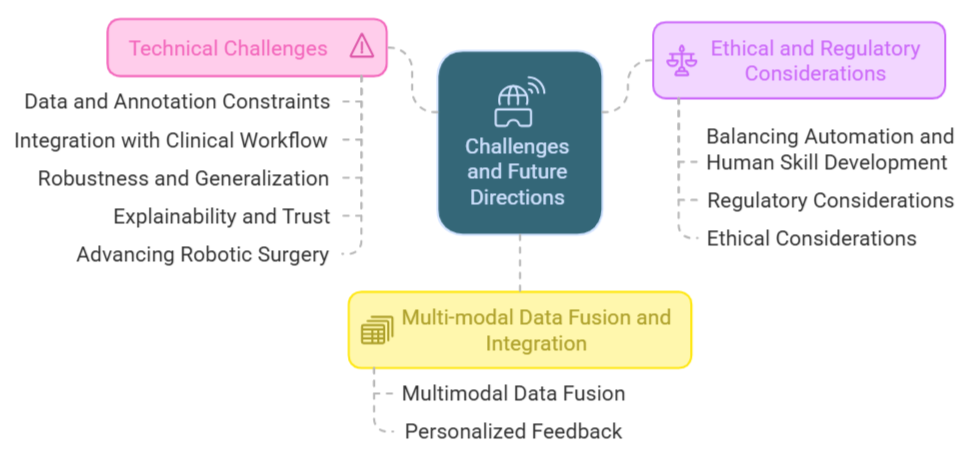}
  \caption{An illustration of potential challenges and future research directions surgical AI development.}
  \label{fig:openissues}
\end{figure}

Surgical scene understanding has made significant strides, yet numerous challenges and open issues persist that require ongoing research and innovative solutions, as shown in Fig.~\ref{fig:openissues}. Addressing these issues is critical for advancing the field and realizing the full potential of AI-driven surgical tools. This section categorizes these challenges into specific areas that are crucial for the development and deployment of surgical AI systems.

\subsection{Technical Challenges}

\subsubsection{Data and Annotation Constraints}
One of the most persistent challenges in surgical video analytics is the scarcity and variability of annotated surgical data. Developing robust semi-supervised \citep{han2024deep}, weakly supervised \citep{khan2024fetr}, and unsupervised learning \citep{stan2024unsupervised} algorithms that can effectively leverage unlabeled data remains an essential area of future research \citep{baradaran2024critical}. Such methods reduce the dependency on large annotated datasets \citep{khan2024ultraweak}, which are costly and time-consuming to produce. Furthermore, fostering collaborations across medical institutions can enable secure sharing of surgical videos and annotations under stringent privacy regulations. These efforts will expand the availability of diverse datasets for training and validating AI models, improving their robustness and generalization.

\subsubsection{Integration with Clinical Workflows}
Although AI models hold great promise for surgical scene understanding, their integration into clinical workflows remains limited due to technical and ergonomic barriers. Future research should prioritize the development of user-friendly interfaces and real-time analysis tools that seamlessly integrate with existing surgical equipment and protocols \citep{byrd2024artificial}. Consideration must also be given to minimizing the ergonomic and cognitive load on surgeons to ensure that AI systems serve as effective tools that enhance, rather than disrupt, surgical performance.

\subsubsection{Robustness and Generalization}
AI models used in surgical scene understanding often struggle with robustness and generalization due to the variability in surgical environments, practices, and equipment across hospitals and geographic regions \citep{matta2024systematic}. Adaptive models that can generalize to new settings without extensive retraining are critical for broader adoption. Promising techniques such as domain adaptation and federated learning offer potential solutions to address these challenges and enhance the reliability of AI models in diverse surgical contexts.

\subsubsection{Explainability and Trust}
The ``black-box'' nature of many deep learning models used in surgery poses significant challenges in terms of explainability and trust \citep{khan2024guardian}. Future research should focus on incorporating explainable AI principles that provide transparent and interpretable insights into AI decision-making processes \citep{samek2019towards}. The enhancement of the explainability of the model will be instrumental in gaining the trust of medical professionals and meeting regulatory requirements, ensuring that AI systems are reliable and safe for use in clinical settings.

\subsubsection{Advancing Robotic Surgery}
AI-driven insights derived from surgical videos have the potential to revolutionize robotic surgery by enabling more intuitive and responsive control systems \citep{bekbolatova2024transformative}. Future research should focus on a deeper integration of AI analytics with robotic systems to support more sophisticated instrument handling, decision making, and task execution. This will improve the precision and safety of robot-assisted surgeries while enabling novel applications in complex procedures.

\subsection{Ethical and Regulatory Considerations}

\subsubsection{Balancing Automation and Human Skill Development}
The increasing integration of AI and robotics in surgery, while enhancing precision and efficiency, raises critical concerns about the atrophy of human skills. As Beane notes in the article ``Today's Robotic Surgery Turns Surgical Trainees into Spectators'' \citep{beane2022today}, surgical trainees are often relegated to passive observers in robotic surgeries, significantly limiting their opportunities to develop essential cognitive and motor skills. This overreliancen automation risks creating a generation of surgeons proficient in operating machines but ill-prepared for handling emergencies or unanticipated complications. To mitigate this, AI systems should prioritize human augmentation by enabling active trainee participation and providing real-time feedback, fostering skill development alongside technological progress \citep{euchner2024designing} \citep{beane2024skill}.

\subsubsection{Ethical Considerations}
As AI becomes increasingly integral to surgical workflows, addressing ethical challenges such as patient safety, data privacy, and bias mitigation will be crucial \citep{o2019legal}. Future research should emphasize the development of robust ethical guidelines and frameworks to ensure that AI systems align with principles of fairness, accountability, and transparency. These guidelines will play a critical role in fostering trust and adoption among clinicians and patients.

From an ethical point of view, the shift towards automation introduces questions of accountability, transparency, and sustainability. As highlighted in The Skill Code \citep{beane2024skill}, surgical AI must be designed to preserve human expertise rather than replace it, ensuring that technology complements the role of the surgeon. Over-standardization and dependency on AI could undermine creativity and adaptability in surgery, diminishing the human element essential for patient-centered care. Addressing these challenges requires a human-centered approach to AI that balances innovation with ethical responsibility, safeguarding both the art and science of surgery.

\subsubsection{Regulatory Considerations}
The rapid advancement of surgical AI requires robust and evolving regulatory frameworks to ensure safety, transparency, and compliance. Existing guidelines, such as AI as a medical device legislation from regulatory bodies such as the FDA and EU MDR, emphasize risk management, performance validation, and post-market surveillance \citep{solaiman2024research}. To align with these standards, surgical AI systems must ensure data security, model interpretability, and clinical efficacy while incorporating mechanisms for continuous learning and adaptation.

Collaboration between AI researchers, medical professionals, and regulatory bodies is essential to establish comprehensive standards that balance innovation with rigorous oversight. These regulations should prioritize patient safety, fairness, and accountability to mitigate biases and unintended consequences. By fostering trust and ethical responsibility, such frameworks will support the sustainable integration of AI into surgical workflows, ensuring both efficacy and safety in clinical applications.
\subsection{Multi-modal Data Fusion and Integration}

\subsubsection{Multi-modal Data Fusion}
Combining information from surgical videos with other types of data, such as patient medical records, real-time sensor data, and histopathological images, can provide a more comprehensive understanding of surgical scenes \citep{xu2024comprehensive}. Multimodal AI systems capable of leveraging this holistic integration will not only improve surgical precision and outcomes, but also enable more personalized and context-aware decision making.

Recent advances in multimodal fusion highlight that integration is no longer limited to simple feature concatenation, but instead relies on specialized strategies for aligning heterogeneous data streams. Broadly, three main categories of fusion approaches are recognized:

\begin{itemize}
    \item \textbf{Early Fusion:} Raw or low-level features from multiple modalities (e.g., pixel embeddings from video frames and embeddings from electronic health records) are concatenated before being fed into a model\citep{baltruvsaitis2018multimodal}. This approach is computationally efficient but often sensitive to noise and modality imbalance.
    
    \item \textbf{Late Fusion:} Independent modality-specific models generate predictions, which are then aggregated through ensemble methods or weighted averaging\citep{yuan2025survey}. In surgery, this could mean combining workflow recognition from video models with complication prediction from patient metadata.
    
    \item \textbf{Hybrid or Joint Representation Fusion:} Modern approaches use deep learning to learn a shared latent space across modalities. Examples include cross-modal attention transformers \citep{tsai2019multimodal}, contrastive learning for aligning video and text embeddings \citep{akbari2021vatt}, and graph-based fusion to jointly model relationships among instruments, anatomy, and patient data \citep{dumyn2024graph}.
\end{itemize}

In the surgical domain, these strategies have been applied to integrate intraoperative video with pathology images for risk stratification \citep{song2025deep}, as well as to combine kinematic sensor data with video for surgical skill assessment and workflow analysis \citep{hu2025vision}. More recently, large vision-language models (LVLMs) have incorporated multimodal adapters to align visual, textual, and sensor features into a unified embedding space, demonstrating improved interpretability and generalization in surgical AI applications \citep{zeng2025surgvlm}. 

\subsubsection{Personalized Feedback}
AI systems have significant potential to enrich surgical training and simulation through personalized feedback mechanisms. Future advancements could focus on the development of virtual reality (VR) and augmented reality (AR) platforms powered by AI to simulate diverse surgical scenarios and complications. These platforms can provide objective assessments and customized feedback to enhance surgeon training skills \citep{guerrero2023advancing}.

\subsubsection{Triplets and Long-Horizon Reasoning}
Triplet recognition expresses surgical activity as an \textit{instrument–verb–target} interaction and offers a compact, clinically meaningful language for events in the operating room. In practice, however, labels are inconsistent across datasets, verbs are vague or applied unevenly, and targets are often annotated at mixed levels of granularity. These inconsistencies reduce clinical interpretability and undermine cross-study comparability. The common evaluation protocols further narrow the lens to short clips or frame-level precision, which hides the real challenges of long procedures such as occlusions, rapid tool exchanges, branching workflows, and step transitions. 

A constructive path forward is to treat triplets as a clinically grounded reporting layer rather than a benchmark target alone. The community should converge on a compact, procedure-aware ontology for verbs and targets with clear synonym maps and explicit tags for safety-relevant interactions. The evaluation should shift from isolated frames to procedure timelines using event-level instances, order consistency with the expected protocol, and lead-time at a specified precision for anticipated events. The newly developed datasets and challenges should include well-defined stress strata, such as multi-tool overlaps, mirror actions, rapid swaps, and prolonged occlusions, and they should contain detailed data cards that document devices, optics, and skill mix so that claims about generalization are testable.


\subsubsection{Beyond Perception: Retrieval and Agents}

Most current systems stop at perception by segmenting instruments, labeling anatomy, or recognizing phases without converting those signals into actionable guidance. Practical operating-room assistance requires retrieving prior cases, guidelines, or safety checklists, answering situation-specific questions, and verifying claims against trustworthy sources such as picture archiving and communication systems (PACS), electronic health records (EHR), and institutional protocols. Vision language models enable these capabilities but introduce risks, including confident errors, mismatched or outdated evidence, and advice that conflicts with local policy. 


Future work should prioritize grounded, \textit{cite-as-you-go} agents that tie every suggestion, whether a next step, a safety reminder, or an explanation, to concrete evidence. Each output should mention timestamps and regions in the current video as well as named external sources, including guidelines and internal checklists. Such an evaluation must extend beyond generic accuracy to measure the impact of the decision, such as error interception and the fraction of each answer that is supported by the retrieved evidence.



\section{Conclusion}
\label{sec:concs}
This survey has highlighted the transformative role of AI, particularly foundation models, in advancing surgical scene understanding, multimodal integration, and decision support. While these developments demonstrate remarkable promise, significant challenges remain. For instance, domain shift across hospitals can result in biased or unsafe predictions, and the high computational complexity of large models introduces latency that undermines real-time intraoperative use. Furthermore, the scarcity of annotated surgical datasets continues to hinder robust generalization, with annotation burdens placing unsustainable demands on clinical experts. Addressing these issues requires targeted strategies, such as federated learning and domain adaptation to mitigate data heterogeneity, pruning and edge optimization to ensure real-time deployment, and semi-supervised or foundation-model-assisted annotation to reduce labeling costs. Looking ahead, future research should focus on robust multimodal fusion frameworks that unify surgical video, patient records, and sensor data for richer context; interpretable AI to enhance clinical trust and accountability; and the exploration of generative digital twin models to improve surgical training and preoperative planning. By explicitly addressing these challenges and pursuing these focused directions, the integration of AI into surgical practice can become safer, more reliable, and clinically impactful.

\bibliographystyle{model2-names.bst}

\bibliography{refs}

\end{document}